\documentclass[12pt]{article}

\usepackage{arxiv}

\usepackage[utf8]{inputenc} 
\usepackage[T1]{fontenc}    
\usepackage{url}     
\usepackage{booktabs}       
\usepackage{amsfonts}       
\usepackage{nicefrac}       
\usepackage{microtype}      
\usepackage{lipsum}		
\usepackage{graphicx}
\usepackage{doi}

\usepackage{amsfonts}
\usepackage{latexsym,amssymb,amsmath,amsthm,amsfonts,graphicx,float}
\usepackage{stix}
\usepackage{tikz}
\allowdisplaybreaks
\usepackage{array}
\usepackage{colortbl}
\usepackage{hhline}
\usepackage{soul}
\usepackage{caption}
\usepackage{subcaption}
\usepackage{verbatim}
\usepackage{amsfonts}
\usepackage{amsmath}
\usepackage{csquotes} 
\usepackage{booktabs}
\usepackage{dsfont}
\usepackage{xparse}
\usepackage{amsmath,amssymb}
\usepackage{graphicx}

\newcommand{\inner}[2]{\left< {#1}, {#2} \right>}

\newcommand{\D}{\Delta^{\varphi}}
\newcommand{\h}{\mathcal{H}^{\D}} 
\newcommand{\hinv}{\Tilde{\mathcal{H}}^{\D}}

\newcommand{\dern}[1]{\frac{\mathrm{d}^{#1}}{\mathrm{d}x^{#1}}}
\usepackage{bbold}

\title{The Hyperdimensional Transform for Distributional Modelling, Regression and Classification}

\date{} 					

\author{\href{https://orcid.org/0000-0003-4564-1672}{Pieter Dewulf$^{\hspace{1mm}\includegraphics[scale=0.06]{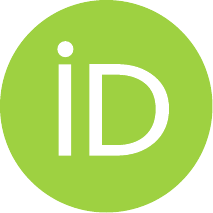}\hspace{1mm}}$},  \href{https://orcid.org/0000-0002-3876-620X}{Bernard De Baets$^{\hspace{1mm}\includegraphics[scale=0.06]{orcid.pdf}\hspace{1mm}}$} , \href{https://orcid.org/0000-0003-0903-6061}{Michiel Stock$^{\hspace{1mm}\includegraphics[scale=0.06]{orcid.pdf}\hspace{1mm}}$}\\
\texttt{pieter.dewulf@ugent.be}, \texttt{ bernard.debaets@ugent.be} , \texttt{ michiel.stock@ugent.be} \\
KERMIT, Department of Data Analysis and Mathematical Modelling \\
Ghent University}

\date{}

\renewcommand{\headeright}{}
\renewcommand{\undertitle}{}

\hypersetup{
pdftitle={The hyperdimensional transform for distributional modelling, regression and classification},
pdfauthor={Pieter Dewulf, Bernard De Baets, Michiel Stock},
pdfkeywords={Hyperdimensional computing, regression, classification, distributional modelling, kernel mean embedding, kernel methods, integral transform, maximum mean discrepancy},
}

\begin{document}
\maketitle

\begin{abstract}
Hyperdimensional computing (HDC) is an increasingly popular computing paradigm with immense potential for future intelligent applications. Although the main ideas already took form in the 1990s, HDC recently gained significant attention, 
especially in the field of machine learning and data science. 
Next to efficiency, interoperability and explainability, HDC offers attractive properties for generalization as it can be seen as an attempt to combine connectionist ideas from neural networks with symbolic aspects.
In recent work, we introduced the hyperdimensional transform, 
revealing deep theoretical foundations for representing functions and distributions as high-dimensional holographic vectors.
Here, we present the power of the hyperdimensional transform to a broad data science audience. 
We use the hyperdimensional transform as a theoretical basis and provide insight into state-of-the-art HDC approaches for machine learning. 
We show how existing algorithms can be modified and how this transform can lead to a novel, well-founded toolbox.
Next to the standard regression and classification tasks of machine learning, our discussion includes various aspects of statistical modelling, such as representation, learning and deconvolving distributions, sampling, Bayesian inference, and uncertainty estimation.
\end{abstract}


\section{Introduction}
\label{sec:introduction}
Hyperdimensional computing (HDC), also known as vector symbolic architectures (VSA), is a computing paradigm based on high-dimensional vectors~\cite{gayler2003vsa_jackendoff, kanerva2009hyperdimensional}. 
Rooted in cognitive science, HDC was proposed in the nineties as a brain-inspired way of computing, as an alternative to conventional von Neumann architectures.
At present, it is being more widely applied with machine learning as the most prominent application domain~\cite{ge2020classification,kleyko2023survey, kleyko2022survey}.
Time and energy efficiency, robustness, interpretability, and multi-modality are benefits that have been extensively demonstrated. 
Many recent applications can be found in the processing of (biomedical) signals~\cite{imani2017voicehd,moin2018emg,schindler2021primer,xu2023hyperspec,zou2022eventhd}, 
biological sequences~\cite{cumbo2020brain,imani2018hdna,kim2020geniehd, shahroodi2022demeter,zou2022biohd}, and robotics~\cite{mitrokhin2019learning, neubert2019introduction}, to name but a few. For more elaborate summaries, we refer to~\cite{ge2020classification,kleyko2023survey,kleyko2022survey}.
Inspired by the human brain, HDC aims at including aspects of artificial intelligence beyond learning, such as memory, analogical reasoning, compositeness, and systematicity~\cite{kleyko2022survey}.
For example, HDC was recently used with neural networks to solve Raven's progressive matrix intelligence tests~\cite{hersche2023neuro}, a problem that eludes computer scientists for ages~\cite{john2003raven,chollet2019measure}. Here, HDC could account for the compositional understanding of the scene and reasoning, yielding new record scores while being orders of magnitude faster than approaches based solely on neural networks.
HDC combines the best of two worlds; while being still a \emph{connectionist} approach like neural networks, using low-level, vectorized representations, it also mimics the symbolic approach of the \emph{good old-fashioned AI}~\cite{kleyko2022survey}.
As a last remark, one could consider HDC again in its infancy~\cite{kleyko2022survey}: it has been gaining momentum over the past years, and further developments may, in the long term, lead to many more powerful applications, such as demonstrated on Raven's progressive matrices.

While HDC has strong conceptual motivations inspired by the human brain, it started as a more empirical area~\cite{kleyko2022survey}.
There are some recent, more theoretical treatises~\cite{kleyko2022survey,thomas2021theoretical}, but mathematical foundations are often lacking in applications, and algorithms are still motivated rather heuristically and intuitively. 
To that end, we recently introduced the hyperdimensional transform as a formal linear integral transform analogous to the ones of Fourier and Laplace. This transform provides foundations for interpreting the \emph{hypervectors}, i.e., the high-dimensional vectors that form the computational units of hyperdimensional computing, as representations of functions and distributions.
In this work, we reintroduce hyperdimensional computing and tailor it toward applications in statistical modelling and machine learning. The overall aims are to provide insight into the current state-of-the-art HDC approaches and to develop well-founded novel approaches.

Concretely, in Section~\ref{sec:background}, we first provide a concise overview of the fundamental concepts of hyperdimensional computing. In Section~\ref{sec:thetransform}, we give an introduction to the hyperdimensional transform and also introduce a new empirical estimation procedure and bipolar approximation. The main body of this work is then divided into three parts: distributional modelling in Section~\ref{sec:distributions}, regression in Section~\ref{sec:regression}, and classification in Section~\ref{sec:classification}. For each part, we summarize the state-of-the-art approaches in HDC and then employ the transform to provide more insight and introduce novel, well-founded approaches.
We discuss the state-of-the-art in classification somewhat more elaborately, as it comprises most of the current work in HDC. However, rather than extensively benchmarking, we aim to provide an intuition of why some approaches might work better than others.

Experimental details and related methods can be found in the supplementary materials. For example, kernel methods such as kernel mean embedding are closely related as they also use high-dimensional representations. Therefore, we consider it essential to distinguish them clearly. Additionally, all code for reproducing the results is available\footnote{\url{https://github.com/padwulf/Chap6_transform_applications}}.

\section{Background on hyperdimensional computing}
\label{sec:background}

\subsection{What is hyperdimensional computing}
\label{sec:whatisHDC}
Any object in hyperdimensional computing is represented as a so-called \emph{hypervector}.
Different types of hypervectors have been described, such as bipolar, binary, ternary, sparse, real-valued, and complex-valued. However, the choice of the specific type is not an essential characteristic of HDC and is usually of secondary importance. We refer to~\cite{schlegel2022comparison} for an extensive overview of the different types. 

A more critical property is that hypervectors represent all kinds of objects with the same fixed dimensionality $D$, ranging from simple objects such as characters and real numbers to more complex ones such as sets, sequences, and graphs.
Regardless of the length of the sequence, for example, the dimensionality of the hyperdimensional representation is fixed at $D$. 
This means that information is not stored locally but is instead distributed \emph{holographically} along the entire vector. Individual components do not stand for specific features of the object but can be seen as random projections based on all features. 
Information is preserved via similarity in the high-dimensional space. 
If the dimensionality is large enough, a second vital property called \emph{hyperdimensionality}, the similarity based on the random projections is assumed to converge. 
Due to the inherent use of randomness, the law of large numbers, and direct access to high-dimensional spaces, hyperdimensional computing naturally gives rise to robust, efficient, and flexible methods for machine learning.

Next to hypervectors, hyperdimensional computing typically involves three operations: \emph{aggregation} (also called addition or superposition), \emph{binding}, and \emph{permutation}. These are used to implement aspects such as reasoning and compositionality. The exact implementation of the operations depends on the type of vector~\cite{schlegel2022comparison}. We illustrate the operations below in Section~\ref{subsubsec:compositeobjects}.
As we will see below, the vectors and operations used in HDC do not necessarily constitute a vector space.

\subsection{Hyperdimensional encoding}

``Similar objects should have similar hyperdimensional representations" is a general guideline for representing data by hypervectors. However, there is usually no universal recipe. Some data types are more straightforward to encode, while others may require some creativity.
In this section, we aim to provide the reader with a flavor of standard practices.

\subsubsection{Atomic objects}
\label{subsucsec:atomicobjects}
The most fundamental atomic objects, such as symbols, characters, or real values, are usually encoded by random sampling of the hypervectors. Objects such as symbols that have no relation to one another can be represented using random i.i.d.\~sampled hypervectors, i.e., all components of the vectors are sampled independently without any notable similarity or correlation.
If the atomic objects are somehow related, for example, if two real values are close, then a higher similarity can be enforced by increasing the number of shared components. 
In practice, one typically subdivides a given interval into discrete levels. The maximum and the minimum levels are represented using random i.i.d.\~sampled vectors. 
In between, the representation of each level is obtained by replacing a fraction of the minimum level with a fraction of the maximum level.
We refer to~\cite{kleyko2022survey} for more detail and discuss the encoding of real values in Section~\ref{subsec:encodinglengthscalel} while introducing a notion of length scale.

\subsubsection{Composite objects}
\label{subsubsec:compositeobjects}
Many objects of higher complexity have been represented by composing representations of atomic ones using the binding, aggregation, and permutation operations already mentioned at the end of Section~\ref{sec:whatisHDC}. Examples are graphs, trees, compounds, sets, sequences, signals, speech, spectra, data records, and images~\cite{imani2017voicehd, imani2018hdna,kanerva2009hyperdimensional, kang2022relhd,kleyko2016holographic,  kleyko2022survey, ma2022molehd,  manabat2019performance, nunes2022graphhd, poduval2022graphd}.
The core idea is that similarity is preserved under the operations, such that compositions of similar atomic objects have similar representations. For example, if two sets or two graphs consist of similar elements,
then their representations should be similar as well.

\paragraph{Binding}
Consider $\phi_x$ and $\phi_y$ the random i.i.d.\~sampled vectors representing the unrelated atomic concepts $x$ and $y$. Binding, denoted by $\otimes$, generates a new hypervector
\[
\phi_z = \phi_x \otimes \phi_y
\,
\]
that represents an association between $x$ and $y$. In the case of real and bipolar hypervectors, binding typically corresponds to elementwise multiplication. 
Assume that $\phi_x$ and $\phi_{x'}$ may be related, but are generated independently from $\phi_y$  and $\phi_{y'}$ (which may be mutually related), then
\begin{equation}
\mathbb{E}\left[ \inner{\phi_x \otimes \phi_y}{\phi_{x'} \otimes \phi_{y'} }\right]
= \mathbb{E}\left[\inner{\phi_x}{\phi_{x'}}\right]
\cdot
\mathbb{E}\left[\inner{\phi_y}{\phi_{y'} }\right]
\,.
\label{lab:tensorproductproperty}
\end{equation}
Here we also assumed that the i.i.d.\ vector representations are zero-centered, i.e., $\mathbb{E}\left[\phi_x\right]=\mathbb{E}\left[\phi_{x'}\right]=\mathbb{E}\left[\phi_y\right]=\mathbb{E}\left[\phi_{y'}\right] = \mathbf{0}$.
This statistical property motivates the use of the symbol $\otimes$ to denote binding, allowing us to interpret binding as a fixed-dimensional approximation of the tensor product. It allows for combining objects of different universes while maintaining a notion of similarity in the joint space via the product.

\paragraph{Aggregation}
Consider a data record $\{(x^i, y^i) \mid i=1,2,\ldots,m\}$ with $m$ keys~$x^i$ and values~$y^i$, typically called \emph{roles} and \emph{fillers}, respectively, in hyperdimensional computing. The entire data record can be represented by aggregating the role-filler bindings, i.e.,
\[
\phi_z = \sum_{i=1}^m \phi_{x^i} \otimes \phi_{y^i}
\,.
\]
For real hypervectors, aggregation corresponds to an elementwise sum. For bipolar hypervectors, a threshold is typically set to retain bipolar vectors, in which case aggregating vectors is no longer associative and one cannot speak of a vector space.

\paragraph{Permutation}
The permutation operation, denoted by $\rho$, generates a new hypervector based on a given one in a deterministic way. The new hypervector should be unrelated to the old one in the sense that $\mathbb{E} \left[ \inner{\phi_x}{\rho(\phi_x)} \right]= 0$. This operation, typically implemented by a single circular shift, allows for encoding positional information. For example, while the association $\phi_x \otimes \phi_y$ is symmetric, $\phi_x \otimes \rho(\phi_y)$ takes the order into account.
This operation is typically used in encoding sequences as aggregations of $k$-mers. For example, for $k=3$, a sequence can be encoded as
\[
\phi_z = \sum_{\textit{ABC } \in \text{ subsequences of } z} \phi_{\textit{A}} \otimes \rho(\phi_{\textit{B}}) \otimes \rho^2(\phi_{\textit{C}})
\,.
\]
This trimer encoding, illustrating the three different operations at the same time, seems simple but works surprisingly well in various sequence  tasks~\cite{najafabadi2016hyperdimensional, imani2018hdna, ma2022molehd, rahimi2016robust}. As we will see in the next section, more complex encodings also exist.

\subsubsection{Black-box methods}
Whereas classical machine learning requires feature engineering and neural networks attempt to learn features, hyperdimensional computing generates a massive number of random features. The idea is that information can then be extracted by simple statistical analyses, for example, by computing similarities that can represent correlations between the components of two vectors. This idea is not exclusive to HDC. 
It is also observed in other techniques such as (physical) reservoir computing~\cite{kleyko2017modality,mcdonald2017reservoir,pieters2022leveraging}, random projection and cellular automata~\cite{kleyko2017modality,yilmaz2015symbolic}, extreme learning machines~\cite{ding2014extreme}, and (deep) randomized neural networks~\cite{gallicchio2020deep}. 
These methods exhibit complex, non-linear behaviours, mapping inputs to high-dimensional representations, from which the labels are inferred using more straightforward techniques like linear regression.
As another example, in~\cite{jiang2022less}, the authors could outperform BERT models for text classification using Gzip for encoding and a simple KNN classifier.
Inspired by such methods, HDC also uses more black-box techniques to map data into hypervectors, ranging from neural networks with convolution and attention mechanism to reservoirs such as cellular automata~\cite{hersche2023neuro,karunaratne2021robust,
mitrokhin2019learning, yilmaz2015symbolic}. 
Using well-known architectures such as convolutional neural networks can allow for better mapping of unstructured data to symbolic hypervector representations. This may improve predictive performance, albeit at the cost of some computational efficiency.

\section{The hyperdimensional transform}
\label{sec:thetransform}
The hyperdimensional transform converts functions and distributions to high-dimensional vectors.
It was extensively presented in our previous work as a formal linear integral transform. In this section, we provide a more concise and intuitive introduction.
For details, we refer to our previous work~\cite{dewulf2023hyperdimensional}.
We refine our view on hyperdimensional encoding, introducing a notion of length scale and normalization.

\subsection[Encoding with length scale l]{Encoding with length scale $l$}
\label{subsec:encodinglengthscalel}
We consider a universe $X$ and a high-dimensional space of the form $S^D$ embedded in $\mathbb{R}^D$, provided with the $D$-dimensional Euclidean inner product $\inner{\cdot}{\cdot}$ that is rescaled by $D$.
A hyperdimensional encoding is defined by a mapping $\varphi:  X \rightarrow S^D$. 
Each component $\varphi_i$ of the vector-valued function $\varphi$ is an outcome of a zero-centered stochastic process, i.e.,
\[
\mathbb{E}\left[ \varphi_i(x)\right] = 0 \quad \text{, for all } x \in X
\,.
\]
We assume the existence of some distance function that quantifies the pairwise similarity between the elements of $X$ and introduce a length scale parameter $l$. If two elements are closer than $l$, the hyperdimensional representations should be correlated, i.e.,
\[
|x-x'|<l \Rightarrow \mathbb{E} \left[\inner{\varphi(x)}{\varphi(x')}\right] > 0 \quad \text{, for all } x,x' \in X
\,.
\]
Conversely, if two elements are farther apart, then the representations should be uncorrelated, i.e.,
\[
|x-x'|\geq l \Rightarrow \mathbb{E} \left[\inner{\varphi(x)}{\varphi(x')}\right] = 0  \quad \text{, for all } x,x' \in X
\,.
\]
These conditions formalize the general principle that similar concepts should have similar representations. Anti-similarity, i.e., $\mathbb{E} \left[\inner{\varphi(x)}{\varphi(x')}\right] < 0$, where representations are more dissimilar than by random chance, is not considered in this work.
The conditions above are given in terms of expected values and can only be guaranteed if the dimensionality $D$ is infinite. For finite dimensionality $D$, some tolerance $\varepsilon$ is assumed.

\begin{figure}
    \centering
    \includegraphics[width=\textwidth]{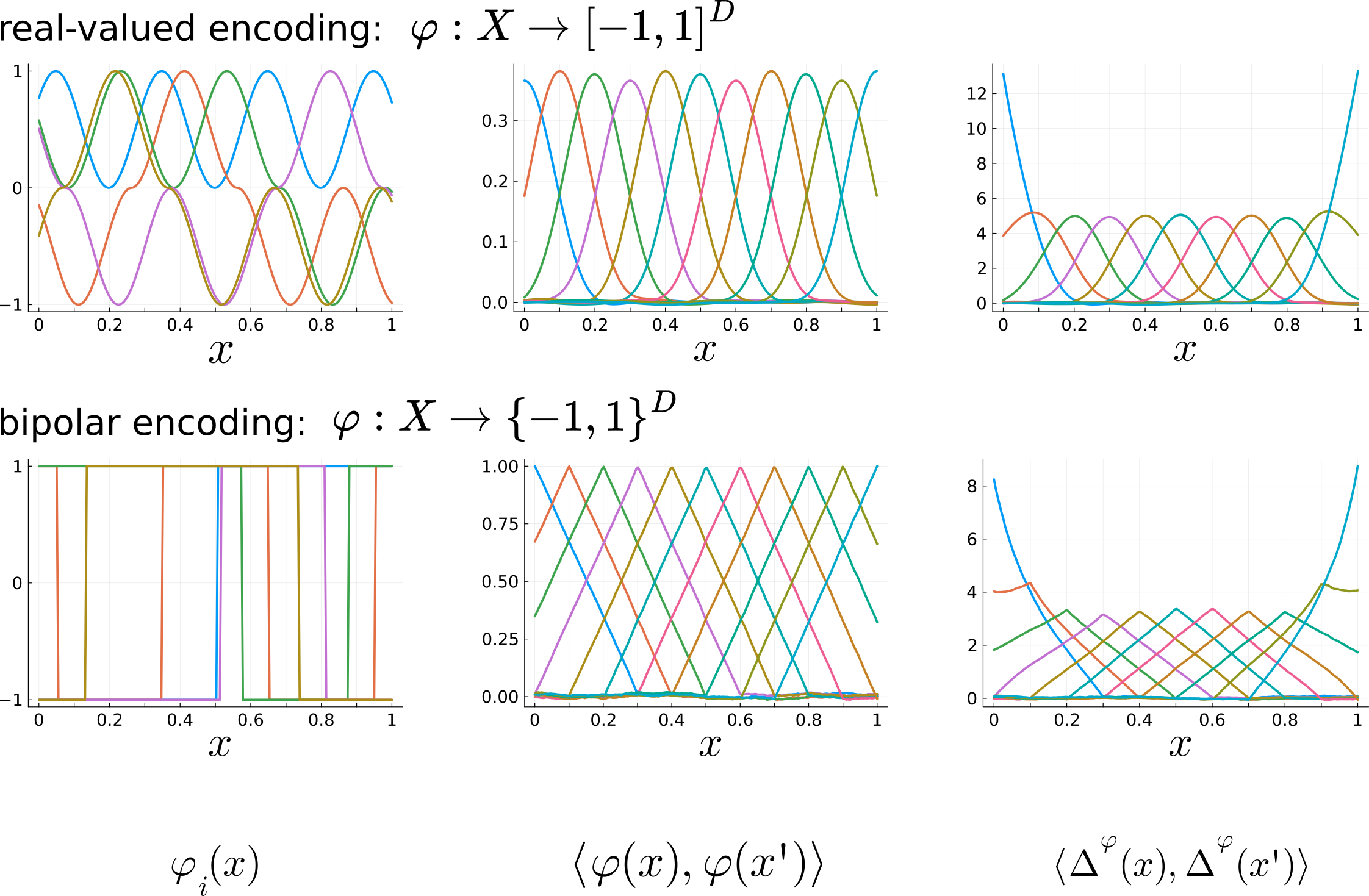}
    \caption{Real-valued (upper)
    and bipolar (lower)
    hyperdimensional encoding with length scale $l=0.3$. 
    The left panels show a few random components. The middle panels show the inner products $\inner{\varphi(x)}{\varphi(x')}$ for fixed $x'$, dropping to zero within the length scale, linearly for bipolar encoding and more quickly for real-valued encoding. 
    The right panels show the inner products $\inner{\D(x)}{\D(x')}$ of the normalized encoding.
    For implementation details, we refer to the supplementary materials, Section SM1.}
    \label{fig:encoding}
\end{figure}

The left panels of Figure~\ref{fig:encoding} visualize examples for encoding real values in the unit interval.

For bipolar encoding, the components of $\varphi$ are random functions alternating between -1 and~1. For real-valued encoding, the components are smooth piecewise functions based on concatenating one-period long segments of cosine functions (with period $l$), each multiplied with a random sign.
The frequency of the random alternation of the functions is determined by the length scale $l$. 
The higher this frequency of random alternation, the faster the correlation $\inner{\varphi(x)}{\varphi(x')}$ drops, shown in the middle panels of Figure~\ref{fig:encoding}. A random phase shift in each component ensures translational invariance.
For the bipolar encoding, we have
\begin{equation}
\mathbb{E} \left[ \inner{\varphi(x)}{\varphi(x')} \right] = \max\left(0, 1 - \frac{|x-x'|}{l} \right)
\,.
\label{eq:triangle}
\end{equation}
A detailed description for constructing $\varphi(x)$ can be found in the supplementary materials, Section SM1.

One can also define a length scale $l$ for other universes $X$ of a more discrete nature. Suppose, for example, a set of unrelated symbols and a distance function that returns 1 for any two of these. Then, i.i.d.\~random sampling corresponds to encoding with a length scale $l=1$. Joining representations with individual length scales via superposition, binding, and permutation operations gives rise to a new joint distance function and length scale. In this way, one can, in principle, analyze any typical HDC encoding technique through its length scale. As an example, for sequences, a length scale can be defined based on the number of shared $k$-mers. For a more detailed example, we refer to the previous on the hyperdimensional transform~\cite{dewulf2023hyperdimensional}.

\subsection{Normalizing the encoding}
We say that a function  $n: X \rightarrow \mathbb{R}$ normalizes\footnote{In our previous work~\cite{dewulf2023hyperdimensional}, we used the expectation to define normalization, i.e.,
\[
 \int_{x'\in X} \frac{\mathbb{E}\left[\inner{\varphi(x)}{\varphi(x')}\right]}{n(x)n(x')} \mathrm{d}\mu(x') = 1 \quad \text{, for all } x \in X
 \,.
\]
However, in practice, the expectation is not always known, and the effective finite-dimensional inner product is a good approximation due to the law of large numbers.}  the hyperdimensional encoding $\varphi: X \rightarrow S^D \subseteq \mathbb{R}^D$ if
\begin{equation}
\int_{x'\in X} \frac{\inner{\varphi(x)}{\varphi(x')}}{n(x)n(x')} \mathrm{d}\mu(x') = 1 \quad \text{, for all } x \in X
\,.
\label{eq:integralnormalization}
 \end{equation}
The normalized hyperdimensional encoding\footnote{When working with bipolar encodings, this division can be implicit such that bipolar hypervectors are retained and efficient bit operations, e.g., for binding or computing inner products can be used.} is then defined as
\[
\D(x) = \varphi(x) / n(x)
\,.
\]
Consider the bipolar encoding in the bottom panels of Figure~\ref{fig:encoding} and assume that $X$ is an unbounded interval. Then, for each $x$, the graph of the function $\inner{\varphi(x)}{\varphi(x')}$ has the shape of a triangle with width $2l$ and height $1$ (Eq.~(\ref{eq:triangle})), and the normalization function is simply the constant $1/\sqrt{l}$. One might compare this to the factor $1/\sqrt{2\pi}$ in the Fourier transform. 
If the universe $X$ is somewhat less homogeneous, for example, when the interval is bounded, then the normalization function $n(x)$ is not a constant and can be interpreted as a function that distorts the space such that every element $x$ has an equal number of neighbors. This is visualized in the right panels in Figure~\ref{fig:encoding}. Each peak has a total mass of 1 
and can be interpreted as an approximation of the Dirac delta distribution with a finite length scale.
While the explicit determination of the normalization function is not always required for machine learning applications, it is needed for a solid theoretical basis for the integral transform.

\paragraph{Computing the normalization function in practice}
Eq.~(\ref{eq:integralnormalization}) corresponds to a Hammerstein equation, which is, in practice, typically solved numerically using an iterative approach~\cite{coclite2000positive,nadir2014numerical}. For a finite-real interval, this takes only a few iterations (see the previous paper on the transform~\cite{dewulf2023hyperdimensional} and the supplementary materials, Section SM1).
For finite sets, e.g., consisting of sequences, the counting measure is used for $\mu$ and Eq.~(\ref{eq:integralnormalization}) takes the form
\[
\frac{1}{|X|} \sum_{x' \in X} \frac{\inner{\varphi(x)}{\varphi(x')}}{n(x)n(x')} = 1 \quad \text{, for all } x \in X
\,.
\]
As illustrated in the supplementary materials, Section~SM1, for a finite set of protein sequences, this is also efficiently solved using the same iterative approach. 
Like with unbounded intervals described above, if $X$ is the more homogenous set of all possible sequences of a certain length without additional constraints, then the normalization function is a constant. In this case, each sequence has the same number of neighbors. It is mainly when $X$ is some realistic subset of all sequences that boundary effects come into play, and solving the equation above might be of interest.

As a last remark, the joint encoding of $X\times Y$, obtained by the binding individual encodings of $X$ and $Y$, is already normalized if the individual ones are. With $x\in X$ and $y\in Y$ and $\D$ and $\Delta^{\psi}$ respective normalized hyperdimensional encodings, we call $\Delta^{\varphi, \psi}(x,y) := \Delta^{\varphi}(x) \otimes \Delta^{\psi}(y)$ a normalized product encoding for the product space $X \times Y$. 

\subsection{Definition of the transform}
\label{subsec:definitionoftransform}
\begin{figure}
    \centering
    \includegraphics[width=1\textwidth]{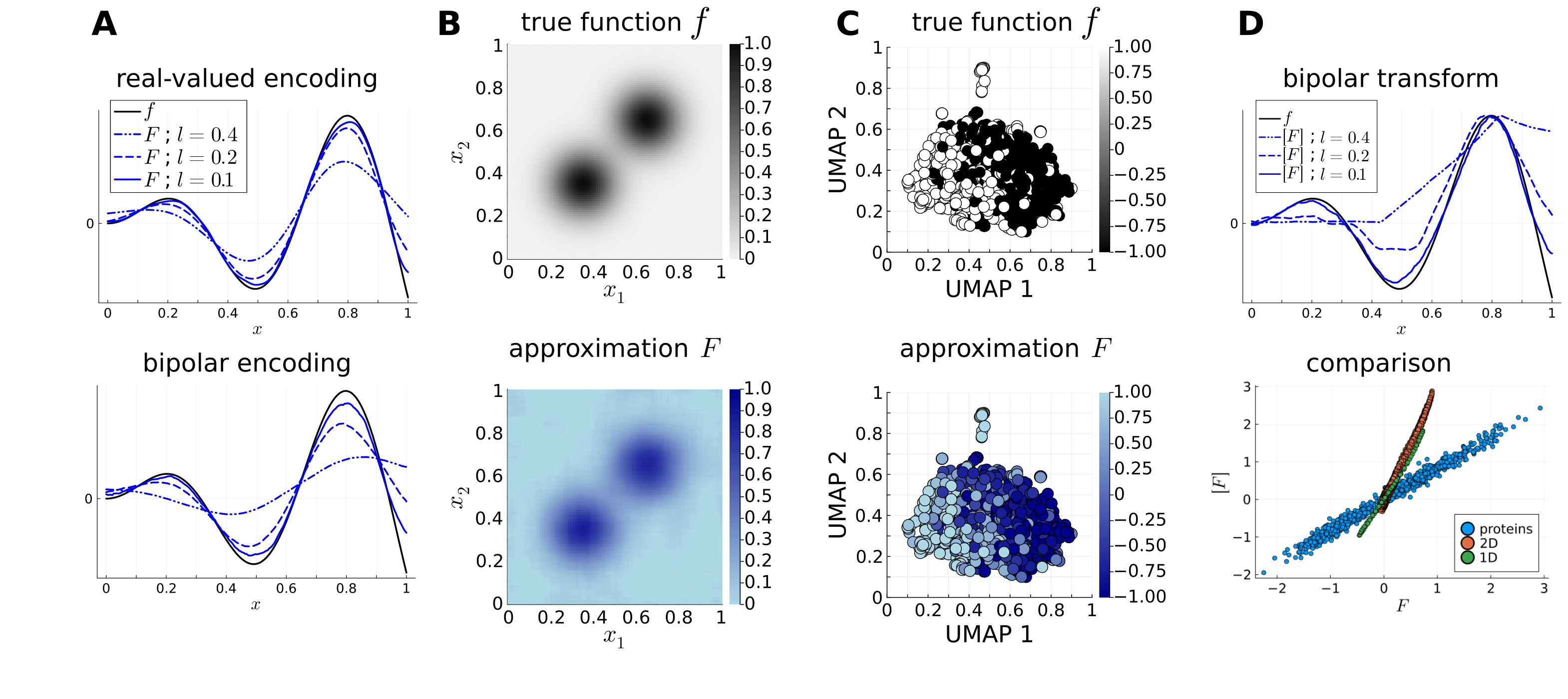}
    \caption{The theoretical transform visualized in one dimension (A), two dimensions (B), and protein sequences representing spaces of higher complexity (C). Subfigure (D) shows the bipolar approximation of the transform, i.e., $[F] = \text{sign}(F)$. The true function and the hyperdimensional representation are visualized in black and blue, respectively. In (A), we varied the length scale $l$ for both bipolar and real-valued encoding. In (B), bipolar encoding was used with $l=0.1$ for both values. In (C), bipolar protein encodings were used with UMAP embeddings for visualization. In (D), bipolar encodings were used, and the bottom panel is based on subfigures (A) with $l=0.1$, (B), and (C). 
    In all cases, the dimensionality of the transform is $D=20\,000$.
     Details on encoding can be found in the supplementary materials, Section~SM1. 
    }
    \label{fig:theoreticaltransform}
\end{figure}
The hyperdimensional transform $F$ 
of a function $f$ w.r.t.~the normalized hyperdimensional encoding $\Delta^{\varphi}$ is defined as
\[
    F = \mathcal{H}^{\Delta^{\varphi}}f := \int_{x\in X} f(x)\Delta^{\varphi}(x)\mathrm{d}\mu(x)    \,.
\]
The hyperdimensional transform $P$ of a probability distribution\footnote{See the supplementary materials, Section~SM5, for a comparison with the kernel mean embedding of distributions.} 
$p$ w.r.t.~the normalized encoding $\Delta^{\varphi}$ is defined as
    \[
    P = \mathcal{H}^{\Delta^{\varphi}}p := \int_{x\in X} \Delta^{\varphi}(x)\mathrm{d}p(x)    \,.
    \]
Here, the Bochner integral with measures $\mu$ and $p$ is used and can be interpreted componentwisely. In general, the transform is an element of $\mathbb{R}^D$. 
Whether $\h$ operates on a function or a distribution should be clear from the context. For simplicity, we do not differentiate notation in this work.
Note that the normalized encoding $\D(x)$ corresponds to the transform of the Dirac delta distribution $\delta_x$, located at $x$, as already suggested by Figure~\ref{fig:encoding}.

The inverse transform of a hypervector $F$ is defined as
\[
     \Tilde{f} = \Tilde{\mathcal{H}}^{\Delta^{\varphi}}F = \inner{F}{\Delta^{\varphi}(\cdot)}
\]
with function evaluation
\[
    \Tilde{f}(x) = \inner{F}{\Delta^{\varphi}(x)}
    \,.
\]
The inverse transform is not the exact inverse operation of the transform but yields a filtered approximation by convoluting with a kernel function $k(x,x')=\inner{\D(x)}{\D(x')}$, i.e., 
\[
(\hinv \h f)(x) = \int_{x' \in X} f(x') \inner{\D(x)}{\D(x')} \mathrm{d}x' 
\,.
\]
The transform is unique (i.e., injective) if the kernel function $k(\cdot,\cdot)$ is strictly positive definite. 
For example, the bipolar encoding with finite length scale $l>0$ from Figure~\ref{fig:encoding} leads to a semi-positive definite kernel function (see Eq.~(\ref{eq:triangle})). In this case, the transform is not necessarily unique. Slightly different functions may lead to the same hyperdimensional transform due to smoothing over length scale $l$. To ensure a strictly unique transform, an arbitrarily small amount of random noise can be added to the encoding, turning the semi-positive kernel into a strictly positive definite one~\cite{dewulf2023hyperdimensional}.

Further, we showed that continuous functions can be approximated arbitrarily close for a finite length scale $l>0$ and that the error decreases as $O(l^2)$ for functions on real intervals~\cite{dewulf2023hyperdimensional}. In practical applications, the length scale $l$ will be treated as a hyperparameter regulating the bias-variance trade-off.
In Figures~\ref{fig:theoreticaltransform}A, B, and C, we visualize some examples. 
In the supplementary materials, Section~SM2, some additional examples are displayed, with much smaller length scales, for example, to encode spectral data.

\paragraph{A bipolar approximation of the transform}
\label{sec:bipolartransform}
Bipolar vectors are equivalent to binary vectors and allow for fast bit operations~\cite{kanerva2009hyperdimensional}. With bipolar encoding, the result of aggregation is typically thresholded to retain bipolar vectors. Similarly, for a bipolar encoding $\varphi: X \rightarrow \{-1,1\}^D$, we define the bipolar approximation\footnote{There is a(n) (infinitesimally) small probability that a component $F_i$ equals~$0$. In this case, it can be assigned randomly 1 or -1 without much impact.} $[F] \in \{-1,1\}^D$ of the transform $F$ as 
\[
[F] = \text{sign}(F)
\,.
\]
Since $\inner{[F]}{[F]}=1$, the approximation $[F]$ can only be correct up to a constant scaling factor. 

The bottom panel of Figure~\ref{fig:theoreticaltransform}D indeed shows a linear relation between the function evaluations of $F$ and those of its approximation $[F]$.
The top panel shows that the bipolar approximation in one dimension is more off for larger length scales $l=0.2$ and $l=0.4$. As with the triangular shape in bipolar encoding, it seems that the bipolar transform estimation yields linear behavior over a scale corresponding to $l$. 
We interpret these results as follows. 
If $x$ and $x'$ are distant enough compared to each other and compared to the boundaries w.r.t.~length scale $l$, then a constant scaling factor $c = c(x) = \frac{\inner{F}{\D(x)}}{\inner{[F]}{\D(x)}}$ could be motivated based on independency and symmetry.
When $x$ and $x'$ are closer than length scale $l$, then their hyperdimensional representations are correlated, and they mix up in $F$, and the scaling factors in $x$ and $x'$ are not necessarily independent.
In conclusion, the smaller the length scale, the better the bipolar approximation.
The scaling factor can be seen as a normalization constant that needs to be calibrated.

\subsection{Integrals and derivatives}
\label{subsection:integralsandderivatives}
The representations of integrals and derivatives in the hyperdimensional space are summarized below. These will be used later on, for example, for marginalizing distributions or implementing physics-based regularization.
With $F$ and $G$ the transforms of $f$ and $g$, inner products can be approximated using 
\[
    \inner{F}{G} 
    =  \int_{x\in X} \tilde{f}(x)g(x)\mathrm{d}\mu(x) 
    =  \int_{x\in X} {f}(x)\tilde{g}(x)\mathrm{d}\mu(x)
    \,.
\]
Here, $\Tilde{f}$ and $\Tilde{g}$ are the inverse transformed functions. 
Equivalently, with $P$ the transform of a distribution $p$, expected values can be approximated using
\[
    \inner{F}{P} 
    =  \int_{x\in X} \tilde{f}(x)\mathrm{d}p(x)
    \,.
\]
Derivatives can be approximated using
\[
\frac{\mathrm{d}^n}{\mathrm{d}x^n} \Tilde{f}(x) = \inner{F}{\frac{\mathrm{d}^n}{\mathrm{d}x^n} \D(x)}
\,.
\]
Depending on the type of the encoding, $\frac{\mathrm{d}}{\mathrm{d}x} \Delta^{\varphi}(x)$ can be computed exactly and efficiently, or via the finite difference method.

\subsection{The empirical transform}
\label{subsec:empiricaltransform}

\begin{figure}
    \centering
    \includegraphics[width=1\textwidth]{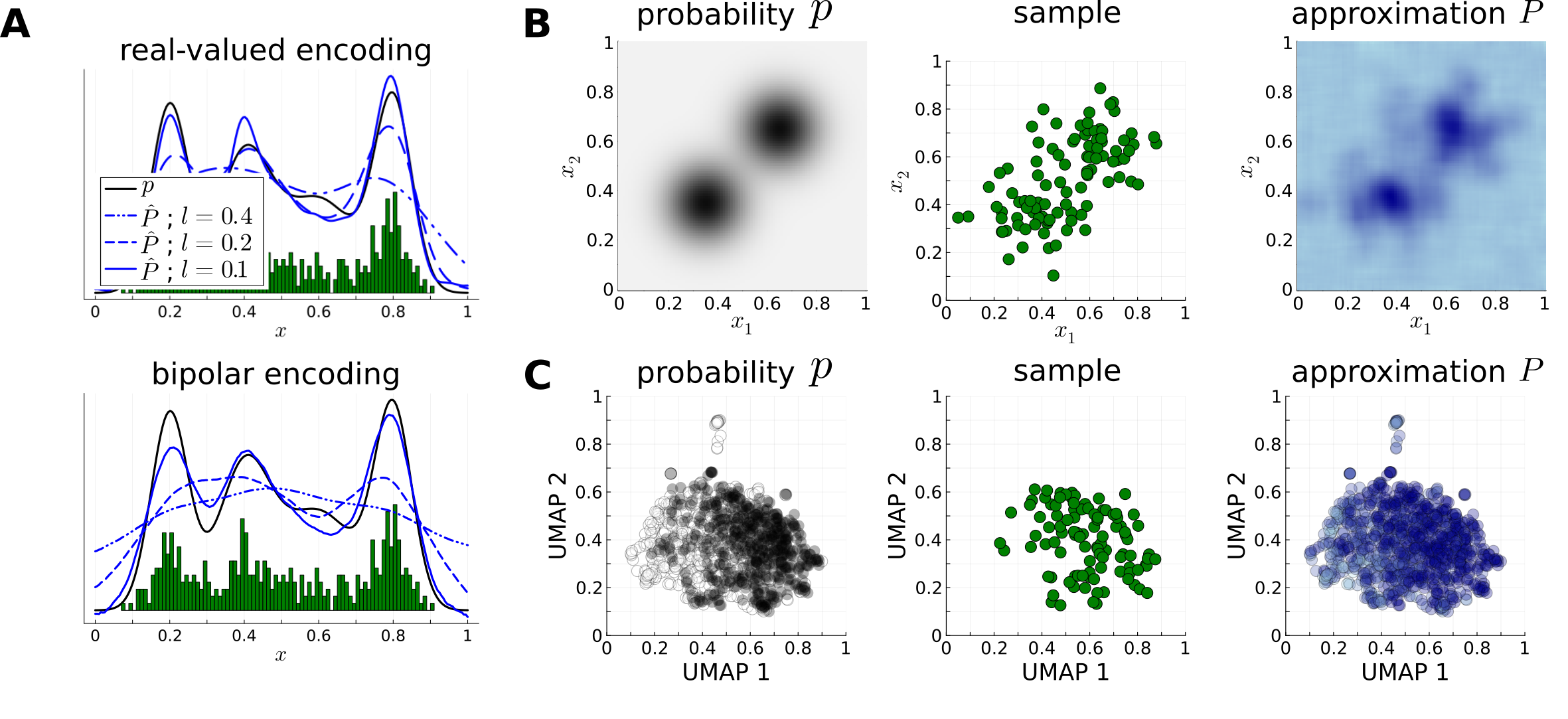}
    \caption{The empirical transform of distributions visualized in one dimension (A), two dimensions (B), and protein sequences representing spaces of higher complexity (C). The true distributions and the hyperdimensional representations are visualized in black and blue, respectively. The data samples, with sizes 400, 100, and 250, respectively, are shown in green. In (A), we varied the length scale $l$ for both bipolar and real-valued encoding. In (B), bipolar encoding was used with $l=0.1$ for both values. In (C), bipolar protein encodings were used with UMAP embeddings for visualization. 
    In all cases, the dimensionality of the transform is $D=20\,000$.
     Details on encoding can be found in the supplementary materials, Section~SM1. 
    }
    \label{fig:empiricaldensitytransform}
\end{figure}

\paragraph{Distributions} The empirical estimate $\hat{P}$ of the tranform $P$ of a distribution $p$ based on a sample $\{x^i \mid i=1,\ldots,m\}$ is given by
\[
\hat{P} = \frac{1}{m} \sum_{i=1}^{m} \D(x^i)
\,.
\]
This simple estimator of the transform is consistent and unbiased. Note that there is a trade-off. 
The transform approximates the true distribution only by a finite precision in $x$ determined by the length scale $l$. 
Smaller length scales allow for a more accurate approximation when more data points are observed, while larger length scales allow for better interpolation when fewer data points are available.
Evaluation of the inverse transform, i.e.,
\[
\hat{p}(x) =\left( \hinv\hat{P}\right)(x) = \inner{\hat{P}}{\D(x)}\,,
\]
allows for density estimation. 
Figure~\ref{fig:empiricaldensitytransform} visualizes some examples.
Note that these density estimations make no assumption on the underlying distribution. The distribution is not parameterized.

Using the normalized encoding, the density estimation is properly normalized:
\arraycolsep=2pt
\begin{eqnarray*}
\int_{x\in X} \hat{p}(x) \mathrm{d}x 
&=& \int_{x\in X} \inner{\hat{P}}{\D(x)} \mathrm{d}x \\
&=& \int_{x\in X} \inner{\frac{1}{m} \sum_{i=1}^{m} \D(x^i)}{\D(x)} \mathrm{d}x\\
&=& \frac{1}{m} \sum_{i=1}^{m} \int_{x\in X} \inner{\D(x^i)}{\D(x)} \mathrm{d}x\\
&\stackrel{(\ref{eq:integralnormalization})}{=}& 1\,.
\end{eqnarray*}
For finite dimensionality, $\inner{\D(x)}{\D(x')} \geq 0$ cannot be strictly ensured, and some tolerance $\varepsilon$ is assumed as a deviation from the expected value. As a result, noise may lead to small negative values for $\hat{p}(x)$ when $\hat{p}(x)$ is small. Small negative probability estimates may be avoided by using $\varphi(x)+\varepsilon$ as an encoding and, in this way, imposing a positive bias on all inner products, which is also a way of regularization. Another approach is post-hoc filtering and replacing small values by zero~\cite{hersche2023neuro}.

\begin{figure}
    \centering
    \includegraphics[width=1\textwidth]{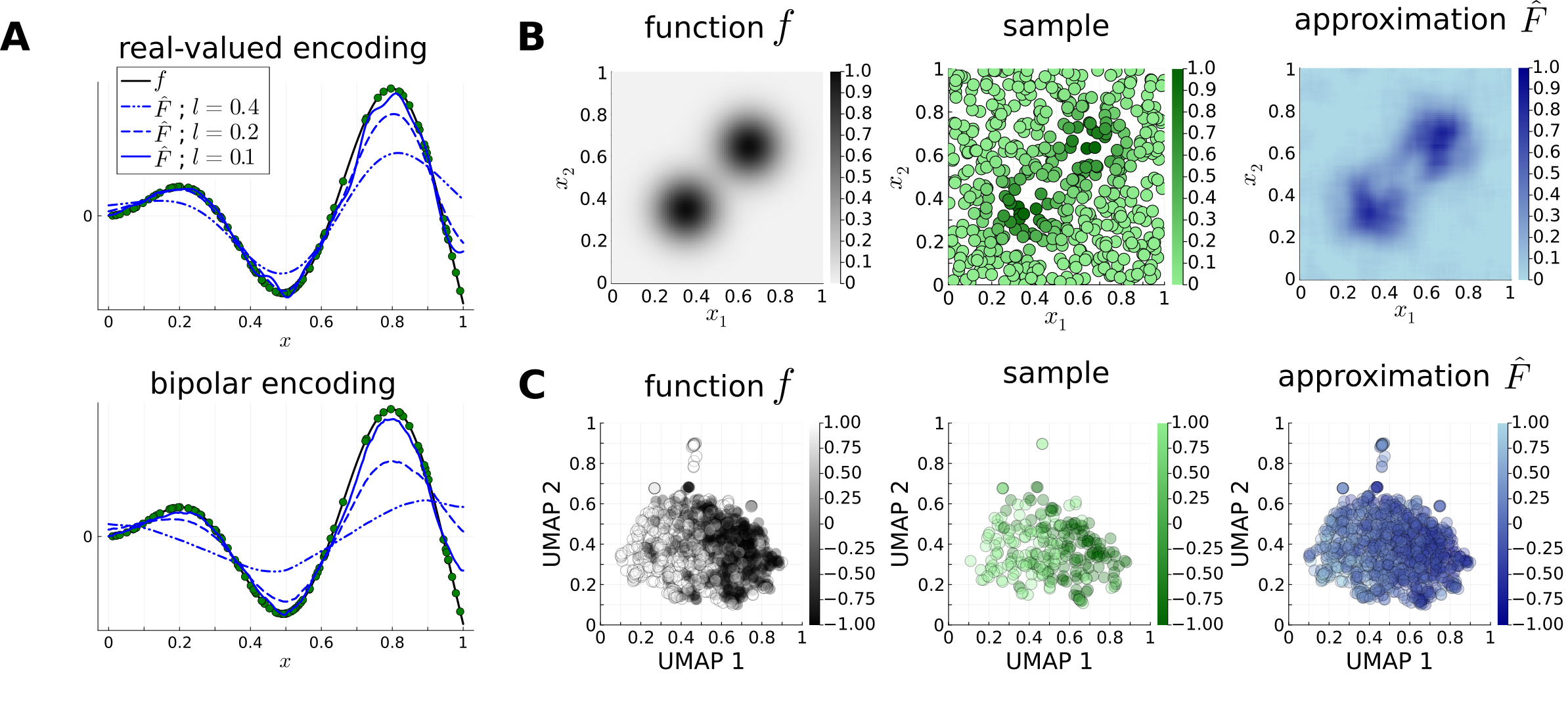}
    \caption{The empirical transform of functions visualized in one dimension (A), two dimensions (B), and protein sequences representing spaces of higher complexity (C). The true function and the hyperdimensional representation are visualized in black and blue, respectively. The data samples, with sizes 50, 500, and 250, respectively, are shown in green. In (A), we varied the length scale $l$ for both bipolar and real-valued encoding. In (B), bipolar encoding was used with $l=0.1$ for both values. In (C), bipolar protein encodings were used with UMAP embeddings for visualization. 
    In all cases, the dimensionality of the transform is $D=20\,000$. Details on encoding can be found in the supplementary materials, Section SM1. 
    }
    \label{fig:empiricalfunctiontransform}
\end{figure}

\paragraph{Functions} The empirical estimate $\hat{F}$ of the transform $F$ of a function $f$ based on a sample $\{(x^i, f(x^i)) \mid i=1,\ldots,m\}$ is given by
\[
\hat{F} = \frac{1}{m} \sum_{i=1}^{m} \frac{f(x^i)}{\hat{p}(x^i)} \Delta(x^i)
\,.
\]
The integral is thus approximated via $m$ point evaluations with volume $1/m$. The denominator $\hat{p}(x^i)$ corrects for varying density of the points $x^i$.
Assuming that the distribution $p(x)$ of the input data is smoother than the variation of the function $f(x)$ and that the length scale $l$ is small enough such that $\hat{p}(x)$ approximates $p(x)$ arbitrarily close, then this estimator is consistent and unbiased. 
For example, if the data points are distributed uniformly, then the density function converges to the uniform distribution for increasing sample size, and $\hat{F}$ converges to $F$. 
Figure~\ref{fig:empiricalfunctiontransform} shows some examples.

For the real line, one can rely on other well-known numerical integration methods such as the trapezium rule. However, the above estimator is more generic and allows for efficient and straightforward estimation for various domain types.

\section{Distributional modelling}
\label{sec:distributions}
In this section, we exploit the hyperdimensional transform to develop well-founded approaches for distributional modelling. Our discussion includes a new distance function for distributions and methods for the 
deconvolution of distributions, sampling, the computation of expectations, conditioning, marginalization, and Bayes' rule. Additionally, we discuss the representation of more complex, joint distributions.
First, we briefly discuss the state-of-the-art in HDC.

\subsection{State-of-the-art HDC approaches for distributional modeling}
Hypervectors have been used implicitly and more explicitly to
model probability distributions. 
For example, in~\cite{thomas2021theoretical}, the authors discuss theoretical foundations for encoding sets of symbols using the superposition operation. This corresponds to representing a uniform distribution of the symbols that make up the set. 
Other works used similarities between hypervectors as class probabilities~\cite{hersche2023neuro, mitrokhin2019learning}. For example, in~\cite{hersche2023neuro}, cosine similarities between hypervectors were pushed through a ReLu activation function and a softmax function to ensure the constraints of a probability distribution. 
To our knowledge, further work on distributional modelling in HDC is somewhat limited. 

\subsection{Maximum mean discrepancy}
We define the maximum mean discrepancy (MMD) for two distributions as the Euclidean distance between their hyperdimensional transforms, i.e., 
\[
\text{MMD}(p,q) = \|P-Q\| = \left\Vert \int_{X} \Delta^{\varphi}(x) \mathrm{d}p(x) - \int_X \Delta^{\varphi}(x) \mathrm{d}q(x) \right\Vert
\,.
\]
The MMD is a distance function for distributions only if the transform is unique, a property discussed in Section~\ref{subsec:definitionoftransform}. If the transform is not unique, then the MMD is only a pseudo-distance function. 
As with the MMD for kernel mean embedding (see supplementary materials, Section~SM5), 
the MMD for the hyperdimensional transform can be used for comparing distributions and statistical testing~\cite{muandet2017kernel}. 
In Table~\ref{tab:proteinmixtures}, the distance is used to compare mixtures of human and yeast proteins. As proteins exhibit a similar structure in the space of all sequences, the differences between the distances are relatively small as compared to the values of the distances. However, we can still distinguish between mixtures containing more human or more yeast proteins.

\begin{table}
    \centering
    \caption{The comparison of distributions illustrated for protein sequences. Five different random mixtures were created out of the 500 human and 500 yeast proteins. The true human and yeast protein distributions are represented by computing the transforms with all available proteins. The dimensionality of the transform is $D=20\,000$. Details on the hyperdimensional encoding of proteins can be found in Section~SM1.}

    \begin{tabular}{l|ccccc}
        & \multicolumn{5}{c}{mixtures} \\
        \hline
        \# human proteins & 250 & 250 & 250 & 125 & 0 \\ 
        \# yeast proteins & 0 & 125 & 250 & 250 & 250 \\
        \hline
        $\text{MMD}(p_{\text{mixture}},p_{\text{human}})$ & 0.911 &  1.03 & 1.15 & 1.49 & 2.05 \\
        $\text{MMD}(p_{\text{mixture}},p_{\text{yeast}})$ & 2.04 & 1.45 & 1.12 & 1.03 & 0.94 \\
        \hline
        estimated $c_{\text{human}}$ & 0.99 & 0.65 & 0.49 & 0.33 & 0.02 \\
        estimated $c_{\text{yeast}}$ & 0.01 & 0.35 & 0.51 & 0.67 & 0.98 \\
    \end{tabular}
    \label{tab:proteinmixtures}
\end{table}

\subsection{Deconvolution}
Distributions can also be analyzed by deconvolution. 
Assume that a distribution $p$ is a convex combination of subdistributions, i.e., 
\[
p = \sum_{i=1}^m c_ip^i 
\,.
\]
Transforming into the hyperdimensional space 
and taking the inner product with $p_j$ yields
\[
\inner{P}{P^j} = \sum_{i=1}^m c_i \inner{P^i}{P^j}
\,.
\]
This simple matrix equation can be solved by inverting the matrix $M$ with components $M_{ij} = \inner{P^i}{P^j}$ representing the correlations between the distributions (see Section~\ref{subsection:integralsandderivatives}).
If the distributions are not linearly dependent, then the problem is well-posed, and the matrix $M$ is full rank.
Typical applications can be found in analyzing experimental data where different signatures contribute to a joint measurement, for example, in single-cell sequencing data or mass spectrometry~\cite{schaffer2019identification,wang2019bulk,xu2023hyperspec}.
As an illustration, in Table~\ref{tab:proteinmixtures}, the hyperdimensional transform is used to deconvolve human and yeast protein mixtures.

\subsection{Sampling}
Using the hyperdimensional transform for sampling is particularly useful when the true distribution is hard to characterize, but a representative sample is available such that the transform can be computed.
For example, there is no parameterization of the distribution of proteins, but one can still sample from that distribution based on its hyperdimensional transform~$P$.
Markov Chain Monte Carlo (MCMC) methods, such as the Metropolis--Hastings algorithm, are typically used in cases with high-dimensional sample spaces, complex distributions, or intractable normalization constants.
Departing from a first sampled element $x$, a new element $x'$ is proposed by slightly changing $x$. The acceptance ratio $r$ is given by 
\[
r= \frac{\inner{P}{\Delta^{\varphi}(x')}}{\inner{P}{\Delta^{\varphi}(x)}}
\,.
\]
If $r\geq 1$, then the proposed element is included in the sample. Otherwise, it is included with 
probability~$r$. In this way, a chain of samples is easily constructed.

In addition, methods such as kernel herding in kernel mean embedding could be used, however, such methods do not provide a random probabilistic sample but rather optimize a representative sample from a given pool~\cite{chensupersamples, muandet2017kernel}.

\subsection{Joint distributions: conditioning, marginalization, and Bayes' rule}
\label{subsec:jointdistr}

\begin{figure}
    \centering
    \includegraphics[width=0.75\textwidth]{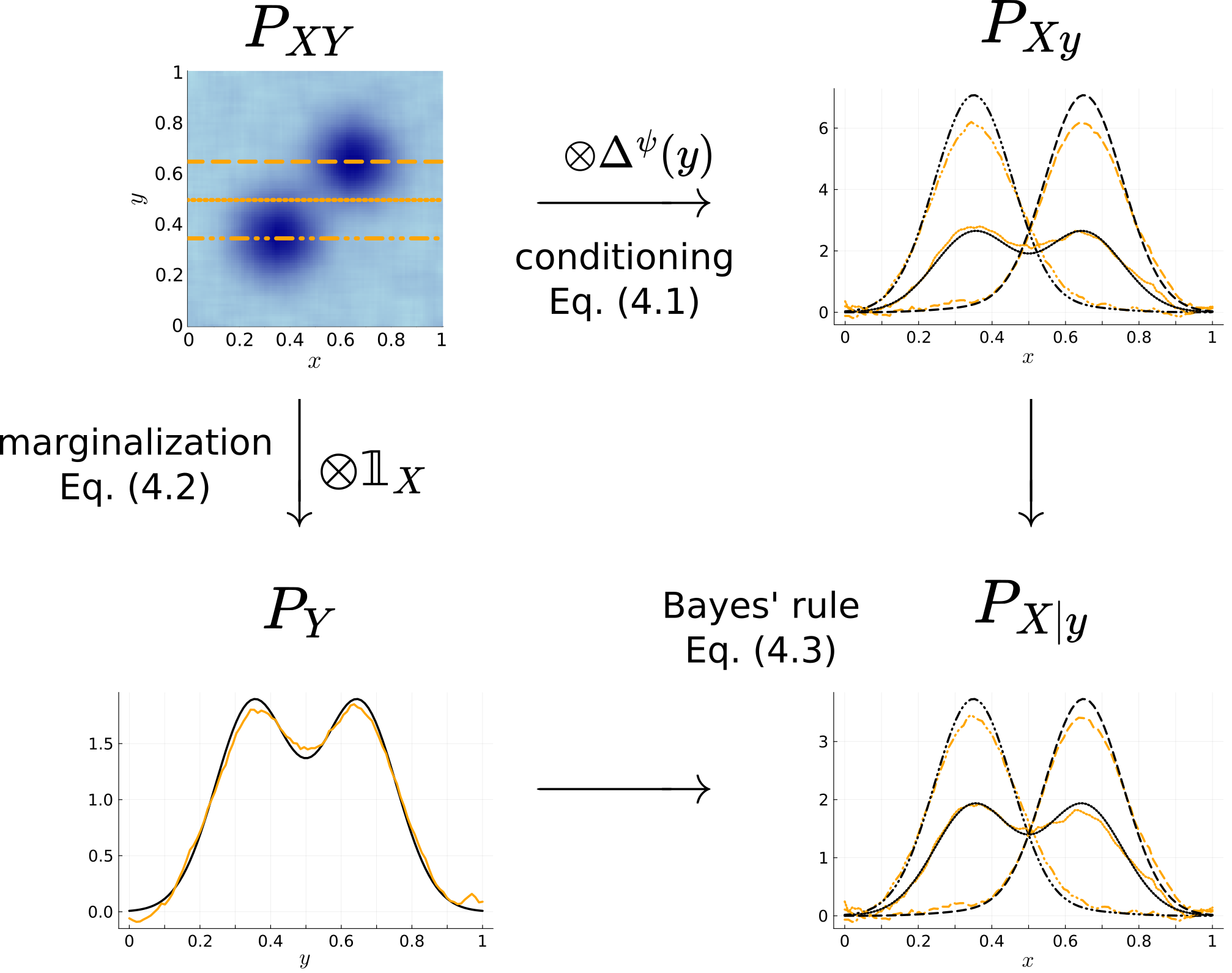}
    \caption{Conditioning, marginalization, and Bayes' rule as operations in hyperdimensional space. The true functions are visualized in black. The hyperdimensional representation of the bivariate distribution is visualized in blue and its conditional and marginal representations in orange. Bipolar encoding with a length scale $l=0.1$ was used for both $X$ and $Y$. The dimensionality of the transform is $D=20\,000$. Details on encoding can be found in the supplementary materials, Section~SM5.}
    \label{fig:marginalization_etc}
\end{figure}

Operations on joint distributions, such as conditioning, marginalization, and Bayes' rule can be explicitly represented in the hyperdimensional space, transforming one hypervector into another.
Consider the product space $X \times Y$ with the normalized product encoding $\Delta^{\varphi, \psi}(x,y) = \Delta^{\varphi}(x) \otimes \Delta^{\psi}(y)$, a bivariate distribution denoted as $p_{XY}$ and its hyperdimensional transform $P_{XY} = \mathcal{H}^{\Delta^{\varphi, \psi}} p_{XY}$ and the following expressions, 
illustrated in Figure~\ref{fig:marginalization_etc}:
\arraycolsep=2pt
\begin{eqnarray}
    P_{Xy} &=& P_{XY} \otimes \Delta^{\psi}(y)\\[.2cm]
    P_Y    &=& P_{XY} \otimes \mathbb{1}_X \\ \label{eq:conditioning} 
    P_{X|y}&=& \frac{P_{XY} \otimes \Delta^{\psi}(y)}{\inner{P_{XY}\otimes \mathbb{1}_X}{\Delta^{\psi}(y)}}\,.
\end{eqnarray}

The first equation represents the conditioning operation on the second random variable at~$y$. Evaluating the conditional hypervector~$P_{Xy}$ 
at~$x$ is equivalent to evaluating $P_{XY}$ at~$(x,y)$:
\arraycolsep=2pt
\begin{eqnarray*}
\inner{P_{Xy}}{\Delta^{\varphi}(x)}
&=& \inner{P_{XY} \otimes \Delta^{\psi}(y)}{\Delta^{\varphi}(x)} \\
&=& \inner{P_{XY}}{\Delta^{\varphi}(x) \otimes \Delta^{\psi}(y)}\\
&=&  \inner{P_{XY}}{\Delta^{\varphi, \psi}(x,y)} \,.
\end{eqnarray*}
Computing $P_{Xy}$ involves smoothing over~$Y$ according to the length scale of the encoder of~$Y$. Therefore, $P_{Xy}$ does not correspond exactly to the transform of the distribution that is first conditioned on~$y$.

The second equation, with $1_X$ the function returning $1$ and its transform $\mathbb{1}_X = \h 1_X$, expresses marginalization. Whereas conditioning is represented by multiplying with a smoothed Dirac delta measure centered at the value to condition on, marginalization uses a function with a constant mass of $1$ everywhere. 
As depicted in Figure~\ref{fig:marginalization_etc}, marginalization results in less smoothing of the highest peaks compared to conditioning; in this case, smoothing is mainly performed over $Y$, as $X$ is marginalized anyway. 

The third equation is obtained by combining the first two and corresponds to Bayes' rule. 
As an alternative formula, the denominator can be written as $\inner{P_{XY}\otimes \Delta^{\psi}(y)}{ \mathbb{1}_X}$, interchanging the order of conditioning on $Y$ and marginalizing $X$.

\subsection{Deep distributional models}

Suppose that $X$ and $Y$ are distributed independently according to individual distributions $p_X$ and $p_Y$ with hyperdimensional transforms $P_X = \h p_X$ and $P_Y = \mathcal{H}^{\Delta^{\psi}}p_Y$. Then, for a large dimensionality $D$, one can write:
\[
P_{XY} = P_X \otimes P_Y
\,.
\]
Indeed, 
\arraycolsep=2pt
\begin{eqnarray*}
    \lim_{D \rightarrow \infty} \inner{P_{XY}}{\Delta^{\varphi, \psi}(x,y)}
    &=& \lim_{D \rightarrow \infty} \inner{P_{X}\otimes P_Y}{\Delta^{\varphi}(x)\otimes \Delta^{\psi}(y)} \\
    &=& \lim_{D \rightarrow \infty} \Big< P_{X}, \D(x) \Big>  \inner{P_Y}{\Delta^{\psi}(y)}
    \,.
\end{eqnarray*}
This result means that representing the joint distribution of two independent variables as a hypervector can be done simply by computing the transform of both distributions individually and binding them. 
Computing the transform and integrating on a product space, which grows exponentially with the number of subspaces of which it is composed, can thus be reduced to integrating on the individual subspaces.

Binding distributions of independent variables and aggregating distributions of the same variable(s) allow for constructing deep probabilistic models, similar to Bayesian networks and graphical models. In particular, these two operations correspond precisely to sum-product networks, which allow for building tractable models from data~\cite{sanchez2021sum}. The hyperdimensional transform offers a way to represent a distribution of high complexity as a single hypervector, such that operations such as marginalization, conditioning, sampling, Bayesian inference, or deconvolution can be performed using straightforward inner products in the hyperdimensional space.
However, the required dimensionality $D$ in relation to the complexity of a network is still an open question.

\section{Regression}
\label{sec:regression}
In this section, we first discuss existing work on regression in HDC, which is, to our knowledge, limited to only three works~\cite{frady2021variable, hernandez2021reghd,mitrokhin2019learning}. Then, we exploit the theoretical foundations of the hyperdimensional transform to present two approaches for regression, one based on estimating the empirical transform and one based on 
empirical risk minimization of the inverse transformed function.
These approaches provide insights into the existing work and provide natural models for regression with attractive properties for uncertainty modelling and regularization.
We consider a training data set $\{(x^i,y^i) \mid i=1,2,\ldots,m\}$.

\subsection{State-of-the-art HDC approaches for regression}
\label{subsec:SOTAregression}
While most of the work in HDC is focused on classification,~\cite{mitrokhin2019learning}~emulates a regression task. 
The authors encoded training input values $x^i$ and output values $y^i$ as hypervectors. The output values $y^i$ were real-valued three-dimensional velocity vectors.
They used a bipolar discrete bin encoding method to represent real values, an idea similar to the one discussed in Section~\ref{subsucsec:atomicobjects}. The representations of the real values were then composed into vectors using record-based encoding as described in Section~\ref{subsubsec:compositeobjects}.
They constructed a hypervector model
\[
M = \sum_{i=1}^m \varphi(x^i) \otimes \psi(y^i)
\,.
\]
A prediction for a datapoint $x$ is then performed by selecting the discrete bin value $y$ such that $\psi(y)$ is most similar to $M \otimes \varphi(x)$. 
Note that, for the bipolar vectors, binding is self-inverse, i.e., $\varphi(x^i) \otimes \varphi(x^i) \otimes \varphi(y^i) = \varphi(y^i)$.
Using the transform, $M \otimes \varphi(x)$ can also be interpreted as the representation of a bivariate distribution conditioned at $x$.

More recent work argues that this approach is an inaccurate method for regression and claims to provide the first, more natural approach for regression using an iterative training procedure~\cite{hernandez2021reghd}. The authors propose a prediction function of the form
\[
\hat{y}(x) = \inner{M}{\varphi(x)}
\,,
\]
with $M$ a hypervector that is trained iteratively. A single pass involves one iteration through all the labels $y^i$, each time applying the update rule
\[
M \leftarrow M + \alpha \left( y^i - \inner{M}{\varphi(x^i)}  \right) \varphi(x^i)
\,.
\]
Here, $\alpha$ is the learning rate, and $y^i - \inner{M}{\varphi(x^i)}$ is the prediction error for data point $i$. The single-pass model was reported to result in a low accuracy because the last training data points of the pass dominate the model. Therefore, multiple passes were preferred. 
The authors also argued that this model is not flexible enough and constructed a multi-model approach. This approach involves a more complex update rule that clusters the inputs and learns $k$ distinct models $M^j$, leading to increased predictive performance.

To obtain a bipolar hypervector model as a lightweight alternative, the authors of~\cite{hernandez2021reghd} used $\hat{y} = \inner{[M^j]}{\varphi(x)}$ as a prediction model for cluster $j$. Here, $[M^j]
\in \{-1,1\}^{D}$ is a bipolar approximation of $M^j$ recomputed each iteration by thresholding at zero. Optimizing $M^j$ for a restricted model $[M^j]$ can be seen as a kind of regularization.

\begin{figure}
    \centering
    \includegraphics[width=0.35\textwidth]{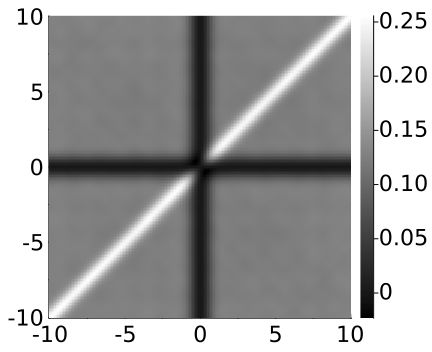}
    \caption{Similarity pattern $\inner{\varphi(x)}{\varphi(x')}$ for the encoding used in~\cite{hernandez2021reghd}.}
    \label{fig:similaritysimcos}
\end{figure}

Further, assuming $x$ is univariate, the real-valued input was encoded using a vector-valued function $\varphi$ with
\[
\varphi_i(x) = \cos(x N_i + U_i) \sin(x N_i)
\,.
\]
Here, $N$ and $U$ are $D$-dimensional vectors filled with elements sampled from a normal distribution and a uniform distribution over $[0,2\pi]$, respectively. 
The emerging similarity pattern $\inner{\varphi(x)}{\varphi(x')}$ is visualized in Figure~\ref{fig:similaritysimcos} for the range of $x$ used in~\cite{hernandez2021reghd}.

\subsection{Empirical transform estimation for regression}
In this section, we employ the empirical transform estimation from Section~\ref{subsec:empiricaltransform} to construct hypervector models for regression, i.e.,
\begin{equation}
    \hat{F} = \frac{1}{m} \sum_{i=1}^{m} \frac{y^i}{\hat{p}(x^i)} \D(x^i)
\label{eq:regressionFhat}
\end{equation}
and
\begin{equation}
    \hat{P}_{XY} = \frac{1}{m} \sum_{i=1}^{m} \D(x^i) \otimes \Delta^{\psi}(y^i)
    \,.
\label{eq:regressionPhat}
\end{equation}
The first hypervector is the empirical transform estimate of the function $f(x)$ and represents a discriminative model with predictions $\hat{y}(x) = \inner{\hat{F}}{\D(x)}$.
The second hypervector is the empirical transform estimate of the joint distribution of inputs and labels. It represents a generative model with predicted probability distributions $\hat{p}(y|x) \propto \inner{\hat{P}_{XY} \otimes \D(x)}{\Delta^{\psi}(y)}$.
The predicted distribution can be normalized as in Eq.~(\ref{eq:conditioning}), or alternatively, empirically in post-processing computations.

\begin{figure}
    \centering
    \includegraphics[width=0.9\textwidth]{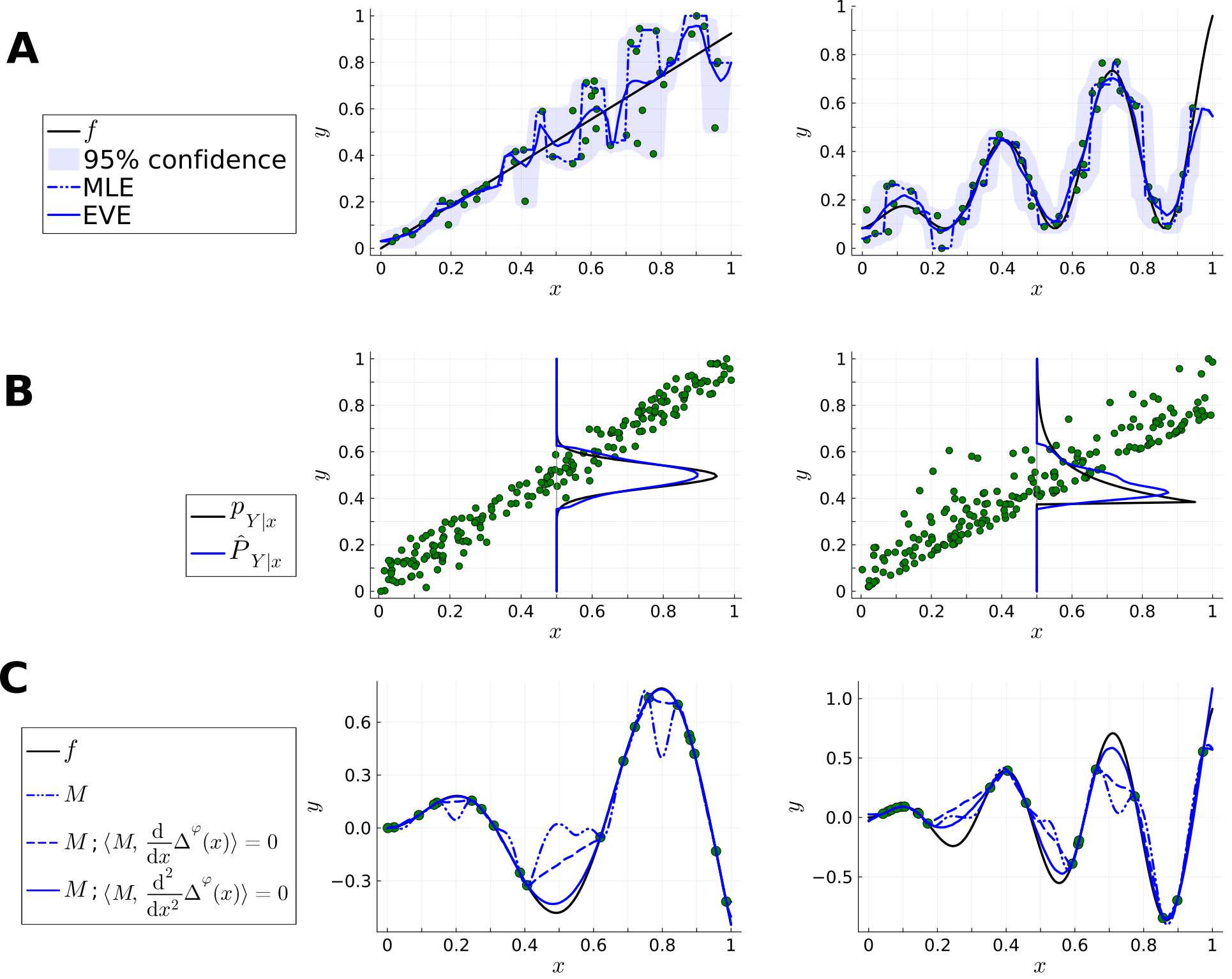}
    \caption{
    Regression using hypervector models via empirical estimation of the transform generatively (A, B) and via empirical risk minimization (C). The true function and the hyperdimensional representation are visualized in black and blue, respectively. The data samples, with sizes 50, 200, and 20, respectively, are shown in green.
    In (A), the predicted distributions are visualized by a 0.95 confidence interval, a maximum likelihood estimate (MLE) and an expected value estimate (EVE). In the left panel, the noise is Gaussian distributed with $\sigma$ proportional to $x$. In the right panel, $\sigma$ is constant. In (B), the predicted distribution for $x=0.5$ is plotted compared to the true noise distribution, which is Gaussian in the left panel and exponential in the right panel. In (C), the hyperdimensional model $M$ obtained by empirical risk minimization is shown, together with physics-based regularized versions. Here, the dimensionality was set to 5\,000 using real-valued encoding with length scale $l=0.1$. In (A) and (B) 20\,000 dimensions with bipolar encoding with $l=0.1$ were used. For more details, we refer to the supplementary materials, Section~SM1.
    }
    \label{fig:regressionwithuncertainty}
\end{figure}

For a visualization of the discriminative approach, we refer back to Figure~\ref{fig:empiricalfunctiontransform}A. The generative approach is visualized in Figure~\ref{fig:regressionwithuncertainty}A, depicting the predicted distribution by the maximum likelihood estimate (MLE),
the expected value estimate (EVE), 
and a symmetric 0.95 confidence interval.
One can see that the MLE is more likely to overfit the observed data, while the EVE is more regularized. 
Based on the transform, we can see that the approach from~\cite{mitrokhin2019learning} explained in Section~\ref{subsec:SOTAregression} can be interpreted as such an MLE estimate. As suggested by the noise dependence in Figure~\ref{fig:regressionwithuncertainty}(a), this MLE estimate may, in combination with the absence of a proper hyperdimensional encoding with a length scale $l$, lead to suboptimal performance.

As depicted in the left panel of Figure~\ref{fig:empiricalfunctiontransform}A, the predicted interval is larger when the labels $y^i$ are noisier. Also, the predicted interval is larger when the trend in the labels $y^i$ is steeper, which can be seen in the right panel. These properties of capturing heteroscedasticity in noise and varying steepness of the actual function are generally not trivial. 
For example, Gaussian processes (see the supplementary materials, Section~SM5) assume a predefined Gaussian noise distribution, and the uncertainty only depends on the locations of the input values $x^i$, without taking the labels $y^i$ into account~\cite{williams2006gaussian}.
Figure~\ref{fig:empiricalfunctiontransform}B emphasizes even more that the predicted distribution for the hyperdimensional model is non-parameterized and that it can learn different types of distributions.

As a last remark, this generative approach is highly suitable for efficient online learning and Bayesian optimization since updating the model requires only adding the new data points, see 
Eq.~(\ref{eq:regressionPhat}).

\subsection{Empirical risk minimization for regression}
\label{subsec:empriskregression}

We observe that the update rule from Section~\ref{subsec:SOTAregression} that was used in~\cite{hernandez2021reghd} can be seen as an implementation of stochastic gradient descent to minimize the loss function
\[
\mathcal{L}(M) 
= \frac{1}{2}\sum_{i=1}^m \left( y^i - \hat{y}^i \right)^2
= \frac{1}{2}\sum_{i=1}^m \left( y^i - \inner{M}{\varphi(x^i)} \right)^2\,,
\]
with predictions $\hat{y}^i = \inner{M}{\varphi(x^i)}$ and gradient
\[
\nabla_M \mathcal{L}(M) = - \sum_{i=1}^m (y^i - \hat{y}^i)\varphi(x^i)
\,.
\]

\subsubsection{Closed-form solution}
\label{subsubsec:regressionclosedformsol}
Instead of relying on stochastic gradient descent, we add an L2 regularization term $\lambda / D \inner{M}{M}$ to the quadratic loss function, ensuring a strictly convex minimization problem with the unique closed-form solution of ridge regression~\cite{mcdonald2009ridge}. The L2 regularization term favors the solution with the smallest norm; see also Section~\ref{subsection:integralsandderivatives} on the correspondence between the L2 norms of hypervectors and functions.
The only difference with ridge regression is that, since our predictions are divided by $D$ due to our definition of the inner product, the solution needs to be multiplied again by a factor $D$. 
Denoting the $m$-dimensional vector $Y = \left[ y^1,\ldots,y^m \right]^\mathrm{T}$ and the $m\times D$ matrix $X = \left[ \D(x^1), \ldots, \D(x^m)  \right]^{\mathrm{T}}$, the solution is given by
\[
M = D(X^{\mathrm{T}} X + \lambda \mathbb{I})^{-1} X^{\mathrm{T}} Y
\,.
\]
Matrix inversion with computational cost $O(D^3)$ is still feasible\footnote{With $A$ a $10\,000 \times 10\,000$ matrix and $Y$ a vector of length $10\,000$, solving $A^{-1} Y$ 
takes approximately 5 seconds on a standard laptop.} for, e.g.,~$D=10\,000$. For larger dimensionalities, one might prefer low-rank, randomized, or gradient-descent-based approximations of the inverse~\cite{halko2011finding}. 
If the data sets are small ($m<D$), then the Woodbury identity~\cite{woodbury1950inverting} allows for exact inversion scaling as $m^3$.
An exciting advantage of the closed-form solution is the algebraic shortcut that allows for leave-one-out cross-validation in constant time, which can be used for efficient model evaluation or hyperparameter tuning~\cite{wahba1990spline, stock2020algebraic, dewulf2021cold}. See 
the supplementary materials, Section~SM3, for more details.

The prediction function $\hat{y}(x) = \inner{M}{\D(x)}$ 
is visualized in Figure~\ref{fig:regressionwithuncertainty}C as model $M$. 
Contrary to the empirical transform estimates in Figure~\ref{fig:empiricalfunctiontransform}A and Figure~\ref{fig:regressionwithuncertainty}A, the function follows the sample data points closely, even using only 5\,000 dimensions. Instead of estimating the transform directly, the inverse transform is explicitly optimized to fit the data.

Due to our length scale $l$ encoding, the model $M$ in Figure~\ref{fig:regressionwithuncertainty}C tends to predict zero when no or few data points are observed. This type of encoding allows modeling the various regions independently and is the basis for approximating any continuous function arbitrarily close when $l>0$. The parameter $l$ can be used to regulate the bias-variance trade-off.
The statement in~\cite{hernandez2021reghd} that an inner product model would not be flexible enough seems, according to our interpretation, rather a consequence of the specific encoding that was used. As shown in Figure~\ref{fig:similaritysimcos}, the encoding has a long-distance correlation pattern, dropping from 0.25 on the smallest distance to only 0.12 on the longest distances. 
As this long-distance correlation pattern lowers the expressivity of a model, it might be the reason why the authors opted for clustering the data, e.g., in 32 clusters, and building independent models.


\subsubsection{Physics-based regularization for out-of-distribution generalization}

A general difficulty for flexible machine learning models such as Gaussian processes and neural networks is that they cannot generalize beyond the training distribution. 
For example, like model $M$ in Figure~\ref{fig:regressionwithuncertainty}C, the prediction of a Gaussian process tends to zero when no data points are observed in an entire kernel bandwidth. 
On the contrary, mechanistic or physics-based models, typically based on differential equations, tend to have better out-of-distribution generalization due to their explicit understanding of underlying principles and constraints. 
The idea of combining the best of both worlds is not new, as briefly discussed in the supplementary materials, Section~SM5 and~\cite{chen2018neural, quaghebeur2022hybrid, solak2002derivative, wang2022physics,  williams2006gaussian}.

As the hyperdimensional transform offers a model $\hat{y}(x) = \inner{M}{\D(x)}$ where function observations and derivative observations are homogeneously represented as inner products (see Section~\ref{subsection:integralsandderivatives}), i.e.,
\[
    \dern{n}\hat{y}(x) = \inner{M}{\dern{n}\D(x)}
    \,,
\]
it is highly suitable for solving differential equations. Consider a linear differential equation of the form
\begin{equation}
    a_0(x)\hat{y}(x) + a_1(x)\frac{{\mathrm d}}{{\mathrm d}x}\hat{y}(x)
    + \cdots + a_n(x)\frac{{\mathrm d}^n}{{\mathrm d}x^n}\hat{y}(x) = c(x)
    \,.
    \label{eq:diffequation}
\end{equation}
Then, an approximate solution to the differential equation as a  hypervector $M$ can be computed via
\begin{equation}
    \inner{a_0(x)\D(x) + a_1(x)\frac{{\mathrm d}}{{\mathrm d}x}\D(x)
    + \cdots + a_n(x)\frac{{\mathrm d}^n}{{\mathrm d}x^n}\D(x)}{M} = c(x)
    \,.
    \label{eq:diffequation2}
\end{equation}
Written as an inner product, the equation corresponds more explicitly to a linear regression problem, with $M$ the vector of model parameters and $c(x)$ the function to fit.
Evaluating this equation at $m$ points, the left operand in the inner product can be written as the $m \times D$ matrix $\Tilde{X}$. As in linear regression, it contains $m$ point evaluations of the encoding $\D(x)$ but now augmented with higher-order derivative observations $\dern{}\D(x)$, $\dern{2}\D(x)$, etc.
The vector $\tilde{Y}$ represents the $m$ point evaluations of the right-hand side of Eq.~(\ref{eq:diffequation2}). The solution for $M$ is then given by
\[
M = D(\tilde{X}^\mathrm{T} \tilde{X} + \lambda \mathbb{I})^{-1} \tilde{X}^\mathrm{T} \Tilde{Y}
\,.
\]
The $m$ points are chosen closer to each other than the length scale $l$ to ensure the differential equation is imposed on the entire domain.

 
By concatenating $X$ and $\Tilde{X}$ and $Y$ and $\Tilde{Y}$, the equations for linear regression and the differential equation can be solved jointly without increasing the $O(D^3)$ complexity of matrix inversion.
In~\cite{dewulf2023hyperdimensional}, we solved differential equations and used regression data points to implement boundary conditions.
In this work, we focus on the regression problem and use the differential equation to implement physics-based regularization. To produce the results in Figure~\ref{fig:regressionwithuncertainty}C, we selected 100 equidistant points to express the differential equation and concatenated them with the regression data points. We solved for $M$ jointly with a small weight on the differential equation points. As a result, the regression data points dominate, and the differential equation is used only where there are no or few regression data points. The interpolations, as constant as possible in the case of $\hat{y}'(x)=0$, and as constant as possible in its derivative in the case of $\hat{y}''(x)=0$, seem more intuitive.
Using the partial derivative expressions from our first work, this can easily be extended for multi-variate problems without increasing the $O(D^3)$ complexity.

\section{Classification}
\label{sec:classification}
In this section, we first discuss existing work on classification. This task has received almost all the attention in machine learning with HDC. Then, analogously to the section on regression, we exploit the theoretical foundations of the transform to present two approaches for classification, one based on estimating the empirical transform, and one based on 
empirical risk minimization of the inverse transformed function.
These approaches provide insights into the existing work and offer natural models for classification with attractive properties for uncertainty modelling and regularization.
We focus on two-class classification problems and consider a training data set $\{(x^i,y^i) \mid i=1,2,\ldots,m\}$ with $y^i \in \{-1,1\}$.

\subsection{State-of-the-art HDC approaches for classification}
\label{subsec:existingHDCclass}
Classification for a data point $x$ is typically performed by constructing class representative vectors $S_+$ and $S_-$ via superposition~\cite{ge2020classification, hernandez2021onlinehd, imani2019adapthd}, i.e., 
\arraycolsep=2pt
\begin{eqnarray*}
S_+ &=& \sum_{\{j \mid y_j=1\}}  \varphi(x_j)\\
S_- &=& \sum_{\{j \mid y_j=-1 \}}  \varphi(x_j)\,.
\end{eqnarray*}
These class representations are then vector-normalized as $[S_+]$ and $[S_-]$. When using bipolar vectors, vector normalization typically corresponds to polarization, i.e., thresholding at zero. When using real-valued vectors, the Euclidean norm is generally used as a scaling factor for normalization.
The predicted class for $x$ is the one of which the normalized representative vector is most similar to $\varphi(x)$, i.e., 
\arraycolsep=2pt
\begin{eqnarray*}
    \hat{y}(x) &=& \text{sign} \left( \inner{[S_+]}{\varphi(x)} - \inner{[S_-]}{\varphi(x)} \right) \\
    &=& \text{sign}\left( \inner{[S_+]-[S_-]}{\D(x)} \right)
    \,.
\end{eqnarray*}
This heuristic approach has been reported as suboptimal~\cite{ge2020classification}.
To obtain higher predictive performance, an iterative algorithm is used, however, at an increased computational cost. 
In a single pass, one iterates through the training data points and applies the following update rule. If $x$ is classified correctly, then do nothing. If $x$ is misclassified positively, then update $S_+ \leftarrow S_+ - \alpha \varphi(x)$ and  $S_- \leftarrow S_- + \alpha \varphi(x)$. Likewise, if $x$ is misclassified negatively, then update $S_+ \leftarrow S_+ + \alpha \varphi(x)$ and  $S_- \leftarrow S_- - \alpha \varphi(x)$. Here, $\alpha$ is a learning rate. To improve the convergence speed, alternative learning rates with iteration or data-point dependencies were proposed~\cite{imani2019adapthd}.
Alternatively, variations exist where $S_+$ and $S_-$ are also updated if $x$ is classified correctly, but the confidence is low, i.e., the similarities of $\varphi(x)$ with $S_+$ and $S_-$ are comparable~\cite{smets2023training}.
As with regression in Section~\ref{subsec:SOTAregression}, optimizing $S_+$ and $S_+$ for restricted bipolar vectors $[S_+]$ and $[S_-]$ in the prediction function $\hat{y}(x)$ can be seen as a kind of regularization.

\subsection{Empirical transform estimation for classification}

The expressions for a discriminative and a generative model for classification are exactly the same as for regression, i.e., 
\begin{equation}
    \hat{F} = \frac{1}{m} \sum_{i=1}^{m} \frac{y^i}{\hat{p}(x^i)} \D(x^i)
\label{eq:classificationFhat}
\end{equation}
and
\begin{equation}
    \hat{P}_{XY} = \frac{1}{m} \sum_{i=1}^{m} \D(x^i) \otimes \Delta^{\psi}(y^i)
\label{eq:classficationPhat}
\,.
\end{equation}
The main difference is that $y$ now takes only values in $\{-1,1\}$. 
For the encoding $\psi(y)$, the values 1 and -1 can be encoded by i.i.d.\ randomly sampled vectors. Normalizing this encoding is trivially a constant as both values have no neighbors.

We are now ready to conduct small experiments and analyze the state-of-the-art approaches. 
We generate two simple synthetic data sets. Assume that the input $x$ is an element of the unit interval and that the positive class is located in the randomly chosen subinterval $[0.145651, 0.3565]$. In the first data set, D1, 100 equidistant sample points $x^i$ are selected, leading to an imbalanced label distribution. The second data set, D2, is perfectly balanced in its labels but is less uniformly distributed: we selected 50, 25, and 25 equidistant data points in the positive and the two negative subintervals, respectively. 
The test set will be just the same as the training set for consistent comparison and to emphasize that classification based on hypervectors is not trivial. The results of different models are given in Table~\ref{tab:classificationresults}. All models are presented as a single vector, and prediction is performed via an inner product with the hyperdimensional encoding. The vector-normalized model vectors between brackets in the table are bipolar approximations.
\begin{table}
    \centering
    \caption{
    The accuracy of HDC classification methods is illustrated on simple synthetic data sets. With the unit interval as the domain, the positive class is located in the randomly chosen interval $[0.145651, 0.3565]$. In the first data set, D1, 100 equidistant sample points are selected, leading to an imbalanced label distribution. The second data set, D2, is more balanced, but less uniformly distributed. We selected 50, 25, and 25 equidistant data points in the positive and the two negative subintervals, respectively. The training and test sets are the same to illustrate very clearly the effect of the models and their difficulties. 
    For the risk minimization model $M$, we used $\lambda'=10^4$. Smaller values for $\lambda'$ lead to the same accuracy for $M$, however, the accuracy in $[M]$ decreases. In all cases, the same bipolar interval encoding $\varphi$ or its normalized version $\D$ is used with $D=20\,000$. For each model, predictions were obtained by taking the inner product. The normalized model vectors between brackets are bipolar.
    }
    \begin{tabular}{c|c|c|c|c|c|c|c|c|c|c}
            & $[S_+]-[S_-]$ & $S_+-S_-$ & $[S_+ - S_-]$ & $\hat{F}$ & $[\hat{F}]$ & $\hat{P}$ & $[\hat{P}]$ & $M$ & $[M]$ \\
        \hline
         D1 & 0.90 & 1    & 0.98 & 1 & 1 & 1 & 1 & 1 & 1\\
         D2 & 0.95 & 0.97 & 0.96 & 0.98 & 0.98 & 0.97 & 0.96 & 1 & 1 \\

    \end{tabular}
    \label{tab:classificationresults}
\end{table}

First, by comparing $[S_+]-[S_-]$ with $S_+-S_-$, it is clear that normalizing the class mean representatives to bipolar hypervectors may severely impact the predictive performance.
One reason is that both classes are normalized independently so that two random approximation errors accumulate. A second reason is that independent normalizations may lead to severe biases. For example, in the case of imbalanced data sets (e.g., D1 in Table~\ref{tab:classificationresults}), based on the hyperdimensional transform of distributions, one can interpret comparing $\inner{[S_+]}{\varphi(x)}$ and $\inner{[S_-]}{\varphi(x)}$ as comparing $p(x|y=1)$ and $p(x|y=-1)$, which implicitly assumes a prior of balanced classes. Also, in case the classes are balanced, biases can occur. Assume, for example, that the negative class is uniformly distributed. In contrast, the positive class is much more locally distributed (e.g., D2 in Table~\ref{tab:classificationresults}). In the very extreme case, the positive class hypervectors align perfectly, and the norm of $S_+$ grows proportionally to the number of data points $m$, while the norm of $S_-$ instead increases proportionally to $\sqrt{m}$. One can see that artificially renormalizing $S_+$ and $S_-$ to the same norm may lead to a substantial bias in predicting the negative class. 
This type of renormalization of $S_+$ and $S_-$ does not necessarily lead to distributions with a total probability of one, as is the case with the hyperdimensional transform, but rather to vectors with the same (Euclidean) norm.
As $S_+-S_-$ is a real-valued model, we also included its bipolar approximation $[S_+-S_-]$ that can be compared more honestly with the model $[S_+]-[S_-]$ based on bipolar vectors.

The main differences between $S_+-S_-$ and the empirical transform $\hat{F}$ in Eq.~(\ref{eq:classificationFhat}) are that $\hat{F}$ uses a normalized encoding $\D(x)$ instead of $\varphi(x)$ and that density corrections are added in the denominator. These somewhat minor numerical modifications improve predictive performance further, which is especially visible in Table~\ref{tab:classificationresults} when comparing the bipolar versions $[S_+-S_-]$ and $[\hat{F}]$.

We observe thus that the standard class mean cluster approach is not optimal for classification, which is (partially) why one must often rely upon iterative training procedures. Better models can be directly constructed based on the empirical hyperdimensional transforms $\hat{F}$ and $\hat{P}$.
The hyperdimensional transform can approximate any continuous function arbitrarily close, which leads to a perfect performance on data set D1.
However, empirically estimating the integral transform is not trivial when the data density varies abruptly, as in data set D2 (237 points per unit length for the positive subinterval and 172 and 39 for the negative ones), leading to a performance that is still suboptimal. See also Section~\ref{subsec:empiricaltransform} on the consistency and bias of the empirical function estimator.
The following section introduces the model $M$ of Table~\ref{tab:classificationresults}, which obtains a perfect performance on both data sets.

\subsection{Empirical risk minimization for classification}
As with regression in Section~\ref{subsec:empriskregression}, we now intend to formulate the iterative algorithm for classification in Section~\ref{subsec:existingHDCclass} as empirical risk minimization.
Using the unnormalized class representative vectors for now, the prediction function can be written as $\hat{y}(x) = \text{sign} \left( \inner{M}{\varphi(x)} \right)$ with $M = S_+ - S_-$ and the update rule can be reformulated as
\[
M \leftarrow M - \alpha (\hat{y}^i - y^i) \varphi(x^i)
\,.
\]
Indeed, if the true class is positive while the negative class is predicted, then $\alpha \varphi(x^i)$ is added twice to $M$, corresponding to one addition to $S_+$ and one subtraction from $S_-$, and vice versa. Recall that we still assume class labels $y^i \in \{-1,1\}$.

If we assume a data-dependent learning rate $\alpha_i =\alpha \inner{M}{\varphi(x^i)}$, then the update rule can be seen as an implementation of stochastic gradient descent of the loss function 
\begin{equation}
\mathcal{L}_{\textit{update}}(M) 
= \frac{1}{2}\sum_{i=1}^m (\hat{y}^i - y^i) \left( \inner{M}{\varphi(x^i)} \right)^2
\,,
\label{eq:updateloss}
\end{equation}
with gradient
\[
\nabla_M \mathcal{L}_{\textit{update}}(M) = \sum_{i=1}^m  (\hat{y}^i - y^i) \inner{M}{\varphi(x^i)} \varphi(x)
\,.
\]
Indeed, in this case the gradient corresponds exactly to the update rule. 
Such learning rates were used to improve the speed of convergence~\cite{imani2019adapthd} for the state-of-the-art classification algorithms.
The loss function above takes value zero if $\inner{M}{\varphi(x)}$ has the correct sign and scales quadratically with $\inner{M}{\varphi(x)}$ otherwise. It can be seen as 
an approximation of the logistic loss, i.e., sigmoid activation combined with cross-entropy, in function of logit $\inner{M}{\varphi(x)}$, depicted in Figure~\ref{fig:loss} and given by
\[
\mathcal{L}_{\textit{logistic}} = \sum_{i=1}^m \frac{1-y^i}{2}\inner{M}{\varphi(x^i)} + \log \left( 1+ \exp(\inner{M}{\varphi(x^i)}) \right)
\,.
\]
Here, $\frac{1-y^i}{2}$ for the bipolar labels $y^i \in \{-1,1\}$ corresponds to $1-y^i$ for standard binary labels in $\{0,1\}$.
The approach of~\cite{smets2023training}, where updates are also performed for correct classification with low confidence, can now be interpreted as shifting the loss functions $\mathcal{L}_\textit{update}$ for the positive and negative labels in Figure~\ref{fig:loss} along the x-axis such that the logistic loss is better approximated. This may result in a better probability calibration and may be the reason for an increased predictive performance.

Again, as a way of regularization to obtain bipolar models, one can optimize $M$ while using its bipolar version for prediction as $\hat{y}(x)=\text{sign} \left( \inner{[M]}{\varphi(x)} \right)$.
If one prefers two cluster representatives, then $\hat{y}(x)=\text{sign} \left( \inner{[S_+]-[S_-]}{\varphi(x)} \right)$ with $S_+=M/2$ and $S_-=-{M}/2$ for the prediction function can be used. Note that only the difference between $[S_+]$ and $[S_-]$ matters, not their exact locations.

\subsubsection{Closed-form solution}
\label{sec:classclosedformsolution}
\begin{figure}
    \centering
    \includegraphics[width=0.6\textwidth]{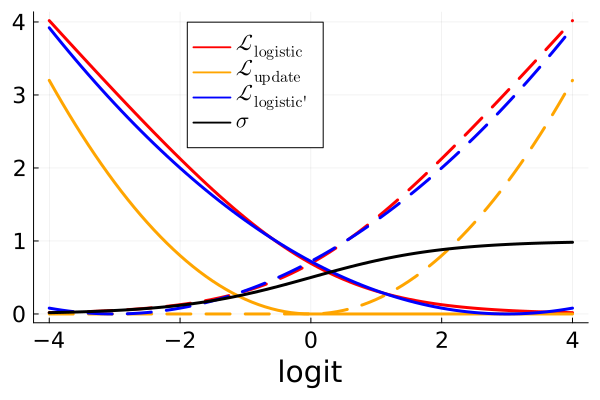}
    \caption{Logistic loss $\mathcal{L}_\textit{logistic}$ and approximations in function of the logit for the positive class (solid lines) and negative class (dashed lines).  The logistic loss in function of the logit is obtained by combining the sigmoid activation function and the binary cross-entropy loss. The approximation $\mathcal{L}_\textit{update}$ is the one from Eq.~(\ref{eq:updateloss}) and $\mathcal{L}_\textit{logistic'}$ is the quadratic approximation described in Section~\ref{sec:classclosedformsolution}.}
    \label{fig:loss}
\end{figure}

Instead of relying on iterative rules with unpredictable convergence, we aim at finding a unique closed-form solution analogous to the regression setting in Section~\ref{subsubsec:regressionclosedformsol}.
We approximate the logistic loss as a parameterized quadratic function, denoted by $\mathcal{L}_\textit{logistic'}$, and add again an L2 regularization term with parameter~$\lambda$. We obtain a solution very similar to the one of standard regularized least squares minimization, i.e.,
\[
    M = a D(X^\mathrm{T} X + \lambda' \mathbb{I})^{-1} X^\mathrm{T} Y
    \,.
\]
However, this solution is multiplied with the constant $a=\beta/2\gamma$ and has a modified regularization parameter $\lambda' = \lambda / 2\gamma$. A derivation is given in the supplementary materials, Section~SM4.
The parameters $\beta$ and $\gamma$ correspond to the distance to the origin of the minimum of the parabola (left of the origin for $y^i=-1$ and right for $y^i=1$) in Figure~\ref{fig:loss} and to the coefficient of the quadratic term.
As depicted in Figure~\ref{fig:loss}, for $\beta=3$ and $\gamma=0.08$, the quadratic loss approximates the logistic loss closely over a logit interval $[-4, 4]$, corresponding to probabilities in $[0.018,0.982]$.

Using a larger $\lambda'=\lambda/2\gamma$ instead of $\lambda$ regularizes the predicted probability closer to this range.
The results of this model~$M$ for our earlier introduced synthetic data sets are included in Table~\ref{tab:classificationresults}. Given that a hypervector can approximate any continuous logit function arbitrarily close, this modified linear regression model for $M$ can, in practice, be more adequate than the empirical transform estimations $\hat{F}$ or $\hat{P}$, especially when the data set is somewhat more difficult, e.g., when exhibiting abrupt density changes as in data set D2. 

As with regression, one can also tune $a$ and $\lambda'$ efficiently using the algebraic shortcut for leave-one-out cross-validation.
This could allow for a better approximation of the logistic loss and improve the calibration of the predicted probability $\hat{p}(x) = \sigma \left( \inner{M}{\D(x)} \right)$. 

Compared to the standard update rules, the closed-form solution of this section, as with regression, offers a unique solution, shortcuts for leave-one-out cross-validation, and possibilities for physics-based regularization. The main drawback might be the computational complexity of $O(D^3)$ for matrix inversion, which is still feasible
for typical dimensions of $D=10\,000$. More efficient approaches for larger dimensionalities were discussed in Section~\ref{subsubsec:regressionclosedformsol}. As a final remark, this section focused on binary classification. The approach can easily be extended for multi-class classification, for example, by binding with the class assigning -1/1 labels for correct and wrong classes or via one-versus-all approaches, and so on.

\section{Conclusion}
\label{sec:conclusions}
Hyperdimensional computing is an area that is rapidly gaining attention in statistical modelling and machine learning.
Some of the reasons are its efficiency, its interpretability, and its symbolic aspects that allow for mimicking reasoning and systematicity.

In this work, we introduced our earlier presented hyperdimensional transform to statistical modelling and machine learning, including new data-oriented concepts, such as the empirical transform.
We employed the transform for distributional modelling, regression, and classification, leading to insights into the state-of-the-art and novel well-founded approaches. New tasks, such as comparing, deconvolving, marginalizing, and conditioning distributions, were tackled. For existing tasks, new aspects were identified for guaranteeing unique solutions, uncertainty quantification, online learning, physics-based regularization, and improving predictive performance in general.

Crucial is our inclusion of a length scale $l$ in the encoding, allowing for a hypervector to approximate any continuous function arbitrarily close. We expect that incorporating this idea, combined with the other findings in this work, will lead to substantial new possibilities for applications with hyperdimensional computing.
Extensive benchmarking of all the newly discussed approaches, ranging from deconvolving distributions and sampling to physics-based regularization and uncertainty quantification, remains future work.

\bibliographystyle{abbrv}
\bibliography{ref.bib}

\end{document}


\section{Encoding}
\label{supp:encoding}
In this section, we describe additional details about encoding. To exclude any ambiguity, all implementation for reproducing the results is available\footnote{\url{https://github.com/padwulf/Chap6_transform_applications}}.

\subsection{Protein encoding}
The data set with 500 human and yeast proteins was encoded using the standard bipolar trimer encoding described in Section~2.2.2. 
We randomly generate $50\,000$ dimensions and select the most ``informative'' $20\,000$, meaning having the largest standard deviation. 
Dimensions with small standard deviations represent more ``general'' features that are more similar throughout the different data points. This selection of informative versus general dimensions is a way of regulating the bias-variance trade-off.

For the normalization function, we used the same iterative procedure as presented in our previous paper~\cite{dewulf2023hyperdimensional}. However, small negative values for finite dimensionality $D$ led to an unstable solution where the normalization takes negative values. Therefore, we used $\varphi(x)+\varepsilon$ with $\varepsilon=0.04$ as an encoding. In this way, the solution of the Hammerstein equation converges quickly. After 15 iterations, the integral for various $x$ has a mean value of $1+4\cdot10^{-6}$ with a standard deviation of $4\cdot10^{-6}$.

\subsection{Interval encoding}
For real-valued encoding, we construct functions
\[
\varphi_{ij}(x') = r_{ij} \frac{1 - \cos(\frac{2\pi}{l} (x'-\theta_i))}{2} \quad \text{ for } x'\in [0,l]
\,.
\]
The functions $\varphi_{ij}$ for various $j$, each with domain $[0,l]$, are then piecewise concatenated to construct a function $\varphi_i$ on $[0,1]$. In each piece, $r_{ij}$ takes a random value in $\{-1,1\}$. In each function, $\theta_i$ is a random phase, allowing for translation invariance for inner products.
Then, the iterative algorithm to compute the normalization function converges quickly~\cite{dewulf2023hyperdimensional}. We approximated it numerically on 200 points in the unit interval and used linear interpolation afterward.

\paragraph{Bipolar interval encoding}
For bipolar encoding, we use the signs of the previous, real-valued encoding, i.e., $\varphi_{ij} = r_{ij}$. The theoretical expected value of the inner products can be analytically expressed as 
\[
\mathbb{E}\left[ \inner{\varphi(x)}{\varphi(x')}\right] = \max\left(0, 1 - \frac{|x-x'|} {l} \right)
\,.
\]
As random switching from 1 to -1 and vice versa occurs uniformly within the length scale, the inner products drop linearly. The normalization function is computed analogously to the normalization of real-valued encoding.

\section{Spectra}
\begin{figure}
    \centering
    \includegraphics[width=0.4\textwidth]{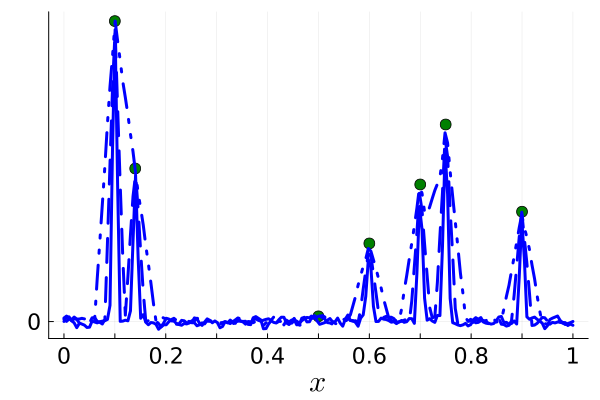}
    \includegraphics[width=0.4\textwidth]{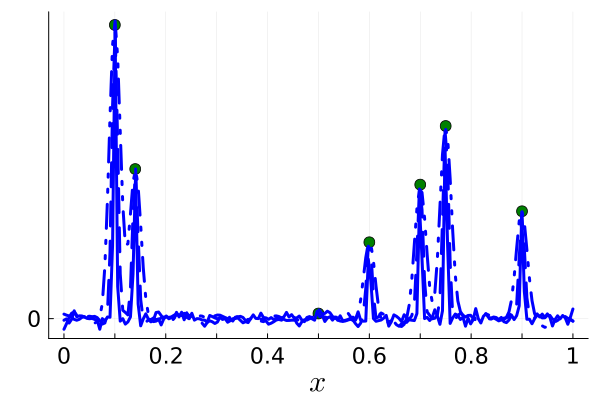}
    \caption{Representing spectra using bipolar encoding (left) and real-valued encoding (right) with various length scales $l=0.04, 0.02, 0.01$. The green dots represent the spectral peaks, and the blue lines represent the hyperdimensional representation. Bipolar encoding generates triangular peaks, while the correlation drops faster for the real-valued encoding.}
    \label{fig:spectra}
\end{figure}

Encoding a spectrum is illustrated in Figure~\ref{fig:spectra} for various length scales.

\section{Leave-one-out cross-validation shortcut and efficient hyperparameter tuning}
\phantom \\
\noindent With a model 
\[
M = D(X^\mathrm{T} X + \lambda \mathbb{I})^{-1} X^\mathrm{T} Y
\]
and prediction function $\hat{y}(x) = \inner{M}{\D(x)}$, the predicted labels $\hat{y}_i$ for the training inputs $x_i$ are given by
\[
\hat{Y} = X (X^\mathrm{T} X + \lambda \mathbb{I})^{-1} X^\mathrm{T} Y =  H Y
\,,
\]
with $H$ the hat matrix, transforming vector of observed labels $Y$ into the vector of predicted labels $\hat{Y}$. Note that, denoting matrix diagonalization as $A=P D P^{-1}$ and using $(PDP^{-1} + \lambda\mathbb{I})^{-1} = P(D+\lambda\mathbb{I})^{-1}P^{-1}$, matrix diagonalization needs to be performed only once to compute the inverse for various $\lambda$.  To perform leave-one-out estimation, the leave-one-out hat matrix $O$ is used
\[
\hat{Y}_{\text{loo}} = OY
\,.
\]
Here, $O$ is the matrix obtained by removing the diagonal elements of $H$ and rescaling each row, i.e.
\[
O_{ij} = \frac{H_{ij}(1-\delta_{ij}H_{ij}) }{1 - H_{ii}}
\,.
\]
We refer to~\cite{dewulf2021cold, wahba1990spline, stock2020algebraic} for more details.

\section{Quadratic approximation of the logistic loss function}
\label{sup:quadraticapproxlog}
We approximate the logistic loss if function of logit $\inner{M}{\varphi(x_i)}$ as a parameterized quadratic function 
\[
\mathscr{L}_{\textit{logistic'}}(M) = \sum_{i=1}^m \frac{1-y_i}{2}\inner{M}{\varphi(x_i)} + \alpha_i + \beta_i \inner{M}{\varphi(x_i)} + \gamma_i \inner{M}{\varphi(x_i)}^2
\,.
\]
Here, $\alpha_i, \beta_i, \gamma_i$ are the parameters that might be different depending on whether $y_i$ is a positive or a negative label.
We denote the model that includes the rescaling of our inner product 
as $M'=M/D$ and add an L2 regularization term $\lambda M'^\mathrm{T} M'$.
We denote the matrix of size $m \times D$ containing the hyperdimensional encodings of the data points by $X$.
The loss function can then be expressed as
\[
\mathscr{L}_{\textit{logistic',L2}} = -\overline{Y}^\mathrm{T}XM' + m \alpha + \gamma (X{M'})^\mathrm{T}(XM') + \lambda {M'}^\mathrm{T}M'
\,.
\]
Here, we choose the $\alpha = \alpha_i$ and $\gamma = \gamma_i$ identically for each $i$.
The $m$-dimensional vector $\overline{Y}$ contains shifted class labels $(y_i-1)/2 -\beta_i$. We choose $\beta = \beta_+ = 1-\beta_- >0$ such that the positive and negative labels are shifted to a location equally far from the origin, and we can write $\overline{Y} = \beta Y$.
Recall that the labels of $Y$ are bipolar.
Imposing that the gradient w.r.t.~$M$ equals zero yields 
\[
M = {D}(2\gamma X^\mathrm{T}X + \lambda \mathbb{I})^{-1} X^\mathrm{T} \overline{Y}
\,,
\]
or equivalently
\[
M = \frac{\beta}{2\gamma}{D}(X^\mathrm{T}X + \frac{\lambda}{2\gamma} \mathbb{I})^{-1} X^\mathrm{T} Y
\,.
\]

\section{Related methods}
In the main text, the discussion of related methods has been limited to HDC. In our previous work~\cite{dewulf2023hyperdimensional}, we discussed links with other integral transforms, such as the Fourier, Laplace, Fuzzy, etc. In this related methods section,  we also provide a brief comparison with the closely related kernel methods, often applied in data science and statistics~\cite{shawe2004kernel, muandet2017kernel}.
These methods also use high-dimensional Hilbert spaces, although more implicitly.
Throughout this section about kernels, we attribute a different meaning to the function symbol $\varphi(x)$. Whereas in the main text, it served as a hyperdimensional encoding, mapping elements to hyperdimensional vectors, in this section, it maps elements to the high-dimensional Hilbert space considered by kernel methods.
Kernels are often used in classical machine learning algorithms where instances $x,x'  \in X$ are not employed directly but rather via an inner product $\inner{x}{x'}$ that is interpreted as a similarity measure between $x$ and $x'$. Typical examples are the kernel perceptron, support vector machines, kernel principal component analysis, Gaussian processes, kernel ridge regression, and kernel mean embedding~\cite{shawe2004kernel, muandet2017kernel}.

\subsection{High-dimensional kernel representation}
Kernel methods assume a non-linear feature map $\varphi: X \rightarrow \mathscr{H}$ that maps the elements of $X$ into some high-dimensional Hilbert space $\mathscr{H}$. Such a feature map defines a positive definite kernel function 
\[
k(x,x') := \inner{\varphi(x)}{\varphi(x')}
\,,
\]
which is used to replace the original inner product $\inner{x}{x}$. This replacement with a highly non-linear function allows kernel methods to build more flexible and powerful learning algorithms. Any positive definite kernel function $k$ can be written as an inner product in a Hilbert space. That Hilbert space is called the reproducing kernel Hilbert space.
The positive definite kernel function is typically computed directly without explicitly computing the feature maps, which may be intractable. For example, the Gaussian kernel function, given by
\[
k(x,x') = \exp\left( - \frac{\|x-x' \|^2}{2\sigma^2} \right)
\,,
\]
is positive definite and has an infinite-dimensional reproducing kernel Hilbert space. 
Similar to the transform, one property of the reproducing kernel Hilbert space is that function evaluation can be expressed as
\begin{equation}
\label{eq:pointevhilbert}
f(x) = \inner{F}{\varphi(x)}
\,,
\end{equation}
with $F$ the representation of $f$ in the reproducing kernel Hilbert space.  Note, however, that this is a theoretical property and that $F$ and $\varphi(x)$ are not explicitly calculated, especially when they are infinite-dimensional. In practice, only inner products are computed. Given a set of data points $\{x_1,x_2,\ldots, x_m\}$, this means that a point $x$ is represented in dual form as a (high-dimensional) vector 
\[
\varphi^*(x)= \left[k(x,x_1), k(x,x_2),\ldots, k(x,x_m) \right]^\mathrm{T}
\,.
\]
A function $f$ is then represented in dual form by the vector $F^*$ via
\begin{equation}
    f(x) = \frac{1}{m} \sum_{i=1}^m F^*_i k(x,x_i) = \inner{F^*}{\varphi^*(x)}
\,.
\label{eq:7_kerneldual}
\end{equation}
As a consequence, Gram matrices of size $m\times m$  need to be computed to represent all $m$ points, such that kernel methods scale at least as $O(m^2)$, and even $O(m^3)$ when matrix inversion is needed. Of course, approximation methods exist to reduce this problem, e.g., the Nystr{\"o}m method~\cite{drineas2005nystrom}.

The similarity to the hyperdimensional transform is remarkable. The (normalized) hyperdimensional encoding represents elements of an abstract universe $X$ in a high-dimensional Hilbert space $\mathbb{R}^D$. Functions can be represented as vectors in that Hilbert space, and function evaluation is expressed as an inner product.

One crucial difference is that, whereas the hyperdimensional encoding also defines a kernel $\inner{\varphi(x)}{\varphi(x')}$, it has an explicit representation in a finite-dimensional reproducing kernel Hilbert space that may stochastically approximate an infinite-dimensional Hilbert space.
Consequently, one can compute $F$ explicitly as an integral transform. In contrast, for kernel methods, a dual representation $F^*$ in Eq.~(\ref{eq:7_kerneldual}) is typically determined using observations $\{(x_i, f(x_i)) \mid i=1,2,\ldots,n\}$ and some least squares method that optimizes the components of $F^*$, i.e., regression.
The explicit representation of $F$ with a finite dimensionality makes HDC a more accessible tool in the design of algorithms, both conceptually and computationally.

One additional element that we introduced for the hyperdimensional representation is normalization. This notion of normalization can be beneficial, as we will illustrate with kernel mean embedding in the section below.

\subsection{Kernel mean embedding}
Kernel mean embedding can be used to represent complex distributions in a non-parameterized way. We provide a brief summary of the kernel mean embedding such that it can be compared to the hyperdimensional transform of distributions. For a comprehensive treatment, we refer to~\cite{muandet2017kernel}.
The kernel mean embedding $P$ of a distribution $p$ is defined as:
\[
P = \int_{x \in X} \varphi(x) \text{d}p(x)
\,,
\]
with $\varphi(x)$ and $P$ both elements of the reproducing kernel Hilbert space. The integral can be interpreted as the Bochner integral.
The definition of the kernel mean embedding has the exact same form as the one of the hyperdimensional transform for distributions. 
However, the kernel mean embedding is defined for distributions and not for functions in $L^2(X)$, which would require a proper normalization, as we will see below.

Again, in practice, $\varphi(x)$ and $P$ are elements of the (infinite-dimensional) reproducing kernel Hilbert space and are not explicitly given.  With a given set of data points $\{x_1,x_2,\ldots, x_m\}$, one relies upon the dual form, i.e.,
\[
P^* = \int_{x \in X} \varphi^*(x) \text{d}p(x) = \int_{x \in X}
\begin{bmatrix}
k(x,x_1) \\[.2cm]
\vdots \\[.2cm]
k(x,x_m) \\[.2cm]
\end{bmatrix}
\text{d}p(x)  
\,,
\]
which can be estimated empirically as
\[
\hat{P}^* = \frac{1}{m} \sum_{i=1}^m
\begin{bmatrix}
k(x_i,x_1) \\[.2cm]
\vdots \\[.2cm]
k(x_i,x_m) \\[.2cm]
\end{bmatrix}
\,.
\]

Whereas the hyperdimensional transform allows for evaluating the distribution directly via the inner product, e.g., $\tilde{p}(x) = \inner{P}{\D(x)}$, this is not possible for the kernel mean embedding since only a dual representation is available. To recover the original distribution from the embedding, a deconvolution problem is solved via constrained optimization. First, a parametrization is proposed of the form
\[
P^* = 
\alpha_1 \begin{bmatrix}
k(a_1,x_1) \\[.2cm]
\vdots \\[.2cm]
k(a_1,x_m) \\[.2cm]
\end{bmatrix} 
+ \ldots 
+ \alpha_n \begin{bmatrix}
k(a_n,x_1) \\[.2cm]
\vdots \\[.2cm]
k(a_n,x_m) \\[.2cm]
\end{bmatrix} 
\,.
\]
Here, $a_1,\ldots,a_n$ are elements in $X$ that are selected by the user and can be interpreted as bin locations in a histogram. The coefficients $\alpha_1, \ldots, \alpha_n$ represent the discrete bin probabilities and can be determined via constrained quadratic programming, with constraints $\alpha_1,\ldots, \alpha_j \geq 0$ and $\sum_{j=1}^n \alpha_j = 1$.
Whereas a proper normalization of the hyperdimensional transform allows for evaluating $\tilde{p}(x) = \inner{P}{\D(x)}$ explicitly, regardless of whether $p$ represents a probability density or a discrete probability, this is not the case for the coefficients $\alpha_1, \ldots, \alpha_n$. The embedding in the kernel Hilbert space has no notion of normalization. Instead, the normalization is enforced via a constrained optimization and depends on the choice of discrete points $a_1,\ldots,a_n$.

In many applications of the kernel mean embedding, for example, in predicting distributions, computing maximum mean discrepancies, sampling from complex distributions, etc., the hyperdimensional transform might be a helpful alternative~\cite{muandet2017kernel, van2020predicting}.
For example, in joint distributions, the hyperdimensional representation of the product space retains the same dimensionality, whereas the dimensionality of the Kronecker kernel representations explodes.
Another advantage is that, as already mentioned, the hyperdimensional transform is intrinsically normalized, and probability evaluation is easily performed via simple inner products. There is no need for parametrization or solving constrained optimization problems to map the embedding to interpretable representations. It allows the hyperdimensional transform to be a bit more generic and also incorporate functions in an analogous way.
However, all these properties come at the cost of noisy behavior when too few dimensions are used.
Future research may compare both approaches in more detail in applications. Not many alternatives exist for this type of problem where the distribution is non-parametrized and complex.

\subsection{Pysics-based regularization}
The idea of combining machine learning models with differential equations is not new, e.g., neural ordinary differential equations, kernel methods with derivative observations~\cite{solak2002derivative, williams2006gaussian, chen2018neural, quaghebeur2022hybrid, wang2022physics}.
For example, in the work of \cite{wang2022physics}, called physics-informed deep kernel learning, differential equations were exploited to increase the performance for scarce data and extrapolation~\cite{wang2022physics}.
Kernel methods and Gaussian processes are highly suitable for differential equations because differentiation is a linear operator and $\mathrm{cov}(\frac{\mathrm{d}}{\mathrm{d}x}y_i,y_j) = \frac{\mathrm{d}}{\mathrm{d}x} \mathrm{cov}(y_i,y_j) = \frac{\mathrm{d}}{\mathrm{d}x} k(x_i, x_j)$~\cite{solak2002derivative}. Consequently, observations and (partial) derivative observations can be combined in one large covariance matrix. 
The main drawback is the computational complexity for matrix inversion, scaling as $O(m^3d^3)$ with $d$ the number of partial derivatives at each data point and $m$ the number of points~\cite{de2021high}. 

For hyperdimensional computing, the explicit representation of derivatives as inner products allows for an approach that retains a complexity of $O({D}^3)$. Also, in the case of multiple variables, partial derivatives and marginalized derivatives can be easily expressed. For example, suppose two variables $x \in X$ and $y \in Y$, and a function that should behave smoothly only in $x$. In that case, using the representation $F\otimes \mathbb{1}_Y$ of the function marginalized on $Y$, allows to impose the partial differential equation ${\partial}f/\partial x = 0$ simultaneously for all $y \in Y$ via $\inner{F\otimes \mathbb{1}_Y}{\der \D(x)} = 0$.
 
\bibliographystyle{abbrv}
\bibliography{ref.bib}